\documentclass[journal]{IEEEtran}
\usepackage{amsmath,amsfonts}
\usepackage{mathtools} 
\usepackage{algorithmic}
\usepackage{algorithm}
\usepackage{subcaption}
\usepackage{multirow}
\usepackage{booktabs}
\usepackage{array}
\usepackage{siunitx}
\usepackage[table]{xcolor}  

\usepackage{graphicx}
\usepackage{listings}
\usepackage{enumitem}
\usepackage{dsfont}

\usepackage{fancyvrb}

\begin{document}
\bstctlcite{IEEEexample:BSTcontrol}

\title{Dialogue Telemetry: Turn-Level Instrumentation for Autonomous Information Gathering}
\author{Dimitris Panagopoulos$^{1}$, Adolfo Perrusqu\'{i}a$^{1}$ and Weisi Guo$^{1}$
\thanks{$^{1}$Faculty of Engineering and Applied Science, Cranfield University, Cranfield MK43 0AL, UK,
        {\tt\small \{d.panagopoulos, adolfo.perrusquia-guzman, weisi.guo\}@cranfield.ac.uk}. 
        This work is funded by EPSRC iCASE with Thales UK (EP/X52475X/1)}%
}

\markboth{Submitted to the IEEE for possible publication. Copyright may be transferred without notice.}%
{Shell \MakeLowercase{\textit{et al.}}: A Sample Article Using IEEEtran.cls for IEEE Journals}



\maketitle

\begin{abstract}
Autonomous systems conducting schema-grounded information-gathering dialogues face an instrumentation gap, lacking turn-level observables for monitoring acquisition efficiency and detecting when questioning becomes unproductive. We introduce \textbf{Dialogue Telemetry (DT)}, a measurement framework that produces two model-agnostic signals after each question--answer exchange: (i) a \textbf{Progress Estimator (PE)} quantifying residual information potential per category (with a bits-based variant), and (ii) a \textbf{Stalling Index (SI)} detecting an observable failure signature characterized by repeated category probing with semantically similar, low-marginal-gain responses. SI flags this pattern without requiring causal diagnosis, supporting monitoring in settings where attributing degradation to specific causes may be impractical. We validate DT in controlled search-and-rescue (SAR)-inspired interviews using large language model (LLM)-based simulations, distinguishing efficient from stalled dialogue traces and illustrating downstream utility by integrating DT signals into a reinforcement learning (RL) policy. Across these settings, DT provides interpretable turn-level instrumentation that improves policy performance when stalling carries operational costs.
\end{abstract}


\section{Introduction}
Autonomous systems increasingly gather mission-critical information through dialogue (e.g., SAR robots interviewing witnesses, disaster coordinators collecting situational reports, and healthcare systems conducting intake). Yet dialogue-based information acquisition remains weakly instrumented compared to other control domains. While navigation exposes pose and velocity observables, and manipulation exposes force/torque sensing, information-gathering dialogues often lack systematic \emph{turn-level} (question--answer exchange) telemetry. As a result, most evaluation is post-hoc at the episode level and provides limited support for online monitoring or closed-loop strategy adaptation \cite{deriu2021survey}.

This instrumentation gap matters because information-gathering dialogues frequently exhibit \emph{diminishing returns}, where repeated probing of the same information categories yields decreasing marginal gains. From a control perspective, this is a throughput degradation problem where the system must detect when a line of questioning becomes unproductive and intervene. In field settings, systems also cannot assume calibrated access to human-state sensing (e.g., reliable affect or fatigue recognition), making \emph{language-derived} observables a practical basis for robust autonomy. In high-stakes information gathering, human interviewers rely on a wealth of implicit feedback (e.g., facial micro-expressions, tonal shifts, and meta-communication) to monitor engagement and detect stalling. Autonomous systems, however, often operate in \emph{signal-sparse} environments where these channels are unavailable or unreliable. Whether due to sensor limitations, remote bandwidth constraints, or the cognitive opacity of the agent itself, these systems lack the implicit feedback loops that regulate human conversation. Consequently, they require \emph{explicit, turn-level telemetry} to serve as a reliable proxy for interaction health, ensuring that mission-critical dialogues remain productive even when social perception is absent.

Search-and-rescue (SAR) operations exemplify the stakes. Coordinators interview witnesses about missing persons under severe time pressure, with the initial intelligence phase described as daunting and chaotic \cite{young2020search, young2023advanced}. These interviews are task-oriented but loosely structured as they are \emph{structured but informal}, typically using prepared question lists (e.g., missing-person questionnaires/interview guidelines) rather than fixed question sequences \cite{young2020search}. Coordinators decide what to probe next while balancing multiple predefined knowledge categories (e.g., last-known location, description, medical needs, companions, etc.). Operational efficiency therefore depends on monitoring, after each exchange, (i) how much useful information remains and (ii) whether the dialogue is drifting into near-repeat probing.

However, current evaluation methodologies exhibit two gaps for schema-driven information gathering.
First, they lack turn-by-turn quantification of \emph{residual} information within predefined categories. Second, they do not reliably detect \emph{semantic revisitation} (i.e., when superficially different questions probe the same conceptual ground), increasing redundancy and wasting interaction time \cite{sacks1974simplest, purver2004theory}. Prior work often focuses on \emph{pre-turn} question selection (e.g., expected information gain) \cite{rao2018learning, aliannejadi2019asking, aliannejadi2021building} or \emph{post-hoc} breakdown detection \cite{higashinaka2016dialogue}. In contrast, systematic \emph{after-each-turn} telemetry that both quantifies remaining learning opportunities and flags revisitation remains understudied \cite{deriu2021survey}.

\textbf{Statement / question.} We ask: \emph{What turn-level observables can instrument schema-grounded information-gathering dialogues so that progress and stalling can be monitored online and used by downstream controllers without requiring causal diagnosis of why degradation occurs?}

\begin{figure}[!t]
\centering
\includegraphics[width=\columnwidth]{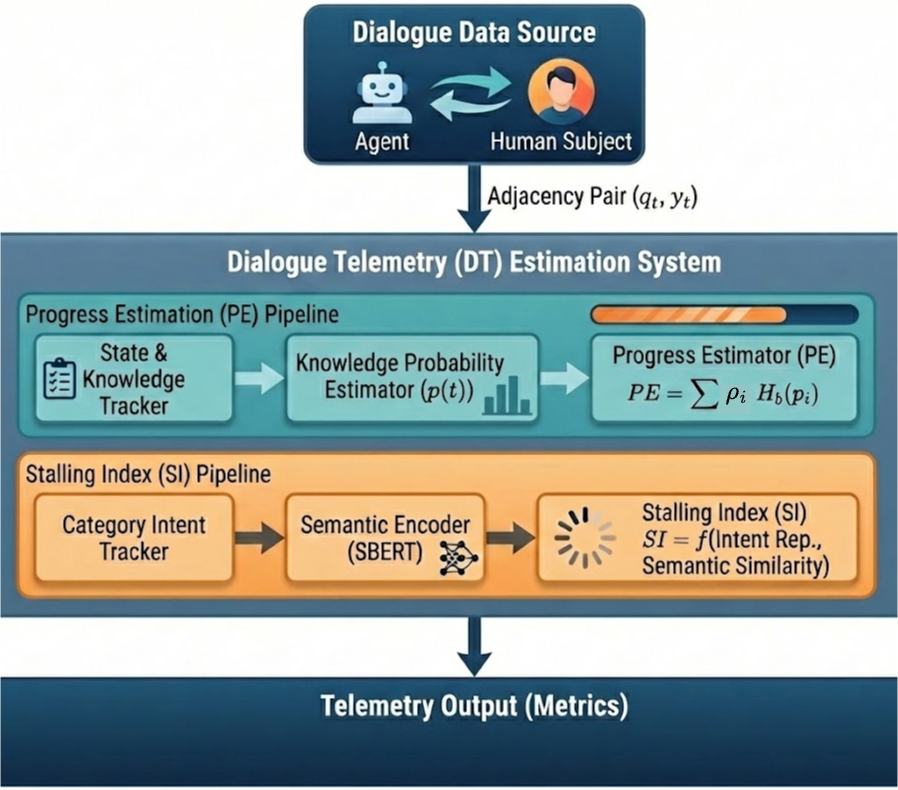}
\caption{The Dialogue Telemetry (DT) measurement architecture. The system ingests turn-level dialogue exchanges (\(q_t, y_t\)) and computes two parallel telemetry signals. The Progress Estimator (PE, blue) tracks the knowledge state per category to quantify residual information potential (including a bits-based variant). The Stalling Index (SI, orange) flags revisitation by combining discrete repetition with semantic embedding similarity. These turn-level observables quantify conversational efficiency and detect stalling online, independent of the downstream control policy.}
\label{fig1:framework}
\end{figure}

We propose \textbf{Dialogue Telemetry (DT)}, a measurement framework that yields two complementary signals \emph{after each turn} (question--answer exchange; adjacency pair), as illustrated in Fig.~\ref{fig1:framework}. A \textbf{Progress Estimator (PE)} operationalizes information-gap reasoning \cite{loewenstein1994psychology} to quantify \emph{residual information potential} per knowledge category. We instantiate PE in two forms: \emph{(a)} a heuristic expected-gain variant combining informativeness rate, residual completeness deficit, and semantic saturation, and \emph{(b)} an information-theoretic variant replacing the deficit with binary entropy to yield expected information gain (bits) per category. A \textbf{Stalling Index (SI)} detects throughput degradation via short-window analysis that fuses discrete repetition patterns with semantic similarity~\cite{reimers2019sentence}. When a category is repeatedly probed with semantically similar exchanges and low marginal gain, SI rises, signaling that the current line of questioning has become unproductive. SI monitors this observable interaction pattern without requiring attribution to specific causes, supporting field deployment where causal diagnosis of questioning strategy or human cognitive state may be impractical. In this work, SI is a within-category stalling signal monitoring degradation conditional on repeated probing of a schema category rather than serving as a global response-quality metric.

DT targets task-oriented information-gathering dialogues where knowledge \emph{categories} are predefined but question order is not prescribed. The telemetry signals are model-agnostic and can support both human monitoring (diagnostic readouts) and automated control (policy observations, stall penalties for reward shaping).

In summary, this work contributes: (1) a schema-grounded measurement framework that provides turn-level observables for information-gathering dialogues, positioning DT as control instrumentation for autonomous systems; (2) mathematically grounded definitions of two telemetry signals, a \emph{Progress Estimator} quantifying residual information potential and expected gain, and a \emph{Stalling Index} detecting interaction-health degradation via observable revisitation patterns independent of causal attribution; and (3) empirical validation in controlled SAR-inspired interview simulations, including integration with RL policies, revealing when telemetry signals improve efficiency and when stall penalties aid versus hinder learning.

\section{Related Work}

SAR guidance and practitioner accounts describe witness interviews as structured but not rigidly
scripted, relying on checklists/training and flexible probing rather than fixed question sequences
\cite{young2020search, koester2008lost}. In cooperative dialogue, classic work on turn-taking,
repair, and clarification shows that participants routinely revisit and refine prior content to
maintain mutual understanding \cite{sacks1974simplest, schegloff1977preference, clark1989contributing, purver2004theory}.
Together, these strands motivate our focus on \emph{after-turn} monitoring of residual information
and revisitation in unstructured, time-sensitive interviews.

We organize related work around (i) \emph{pre-turn} question selection and policy optimization,
(ii) \emph{post-hoc} breakdown detection, and (iii) evaluation efforts that motivate the need for
standardized \emph{turn-level} monitoring.

\paragraph{Pre-turn question selection and policy optimization.}
A large body of work chooses the next question \emph{before} asking, often using information-theoretic
objectives such as expected information gain (EIG) or expected value of perfect information (EVPI)
\cite{rao2018learning}. In NLP, models rank clarifying questions for open-domain information-seeking
conversations using utility and information-based criteria \cite{aliannejadi2019asking, aliannejadi2021building}.
Related information-theoretic approaches appear in visually grounded and referential dialogue, where
questioners explicitly maximize information gain or use informative rewards \cite{lee2018answerer, shukla2019should}.
These methods optimize \emph{which question to ask}, but they do not provide a standardized,
\emph{after-turn} way to track cumulative progress by schema category and to detect when the interaction
is degenerating into unproductive revisitation.
Separately, RL-based dialogue policy work studies learning efficiency issues such as getting trapped
in dead-end trajectories and proposes mechanisms to detect and recover from them during exploration
\cite{zhao2024rescue}.

\paragraph{Post-hoc breakdown detection.}
Dialogue breakdown detection identifies inappropriate or unproductive system behavior through
utterance- or turn-level analysis and produces labels intended for diagnosis or evaluation
\cite{higashinaka2016dialogue, terragni2022betold, ng2020improving}. These approaches are typically
post-hoc and largely categorical (e.g., breakdown vs.\ non-breakdown). They may rely on surface cues
or on intent/entity representations rather than explicit semantic-progress telemetry. In contrast,
our SI supplies a \emph{continuous, short-horizon} signal designed for control and monitoring,
targeting \emph{semantic} revisitation even when surface phrasing differs.

\paragraph{Evaluation and turn-level monitoring.}
Surveys highlight fragmentation in dialogue evaluation and the prevalence of global, episode-level
metrics rather than standardized turn-level monitoring \cite{deriu2021survey}. DT contributes two
lightweight, model-agnostic \emph{turn-level} measures: one that estimates residual information
\emph{per category} (PE) and one that measures short-window revisitation (SI), intended to guide the
very next decision and to support interpretable evaluation.

\paragraph{HRI and task-oriented dialogue efficiency.}
HRI work emphasizes the importance of trust, its assessment, and its repair in human--robot interaction
\cite{campagna2025systematic, esterwood2023theory}. In task-oriented robot instruction settings,
information-theoretic clarification has been shown to reduce the number of questions while improving
task success \cite{deits2013clarifying, tellex2012toward}. These systems often embed efficiency
mechanisms within specific architectures. DT instead provides \emph{model-agnostic telemetry} that can
augment diverse systems with explicit turn-level progress visibility and revisitation detection when a
task schema (knowledge categories) is available.

\paragraph{Information-theoretic analysis of long LLM conversations.}
Recent work quantifies diminishing returns in unconstrained multi-turn LLM conversations by modeling
dialogue as an interactive (feedback) communication channel and introducing turn-level efficiency
measures, including \emph{Information Gain per Turn} (IGT), \emph{Token Waste Ratio} (TWR), and an
interactive-channel capacity bound $C_{\text{int}}$ \cite{gorlequantifying}. DT is complementary. Rather than estimating per-turn throughput in unconstrained chat, it provides \emph{schema-grounded}
telemetry for task-oriented information gathering. PE quantifies residual information potential
\emph{per category} (including a bits-based variant), and SI flags semantic revisitation patterns that
signal unproductive probing in checklist-driven interviews. Both frameworks address diminishing
returns in multi-turn dialogue, but DT’s schema grounding yields actionable observables for monitoring
and closed-loop control when information requirements are predefined (e.g., SAR interviews, medical
intake, incident reporting).

\section{Dialogue Telemetry (DT)}
\label{sec3:DT_formulation}

We treat schema-grounded information-gathering dialogues as measurable processes over a predefined task schema.
A \emph{turn} is one question--answer exchange (adjacency pair). We formalize a dialogue as a sequence of $T$ turns
$D=\{(q_1,y_1),\ldots,(q_T,y_T)\}$, where $q_t$ is the question at turn $t$ and $y_t$ is the corresponding answer.
Let $\mathcal{M}$ be the set of knowledge categories (e.g., location, time, description).

At turn $t$, after receiving $y_t$, the system maintains a hybrid state over categories:
\begin{equation}\label{eq1:hybrid_state}
s(t)=\Big\{ \big(\upsilon_i(t),\, e_i(t),\, m_i(t),\, k_i(t)\big)\ \Big|\ i\in\mathcal{M}\Big\}.
\end{equation}
Here $\upsilon_i(t)\in[0,1]$ denotes the estimated completeness for category $i$,
$e_i(t)\in\mathbb{R}^d$ is the running semantic embedding trace for $i$,
$m_i(t)\in\mathbb{N}$ counts how many times $i$ has been queried up to turn $t$,
and $k_i(t)\in\mathbb{N}$ counts how many queries to $i$ produced an informative update.

Let $i_t\in\mathcal{M}$ be the category targeted at turn $t$. The per-turn completeness gain is
$\Delta \upsilon_i(t)=\upsilon_i(t)-\upsilon_i(t-1)$, where $\Delta \upsilon_i(t)=0$ for $i\neq i_t$.
We define an informative update indicator $\mathbb{I}_i(t)=\mathds{1}\{\Delta\upsilon_i(t)>\varepsilon_{\upsilon}\}$
with a small threshold $\varepsilon_{\upsilon}\ge 0$, and update counts by
\begin{align*}
m_i(t)&=m_i(t-1)+\mathds{1}\{i_t=i\},\\
k_i(t)&=k_i(t-1)+\mathbb{I}_i(t).
\end{align*}

For semantic tracing, let $\phi(\cdot)$ map an answer span to an embedding vector. We update the embedding trace as
$e_i(t)=e_i(t-1)+\mathds{1}\{i_t=i\}\,\phi(y_t)$ (sum-trace; other aggregations such as running means are possible).
Candidate next questions are specified by a strategy--category pair, and answers are mapped to per-category gains
$\Delta\upsilon_i(t)$ and embedding increments $\phi(y_t)$.
In practice, $\upsilon_i(t)$ is provided by a domain scoring module (manual rubric, classifier, or LLM-based assessor).
DT is agnostic to this choice and operates on the resulting per-turn gains.

\subsection{Progress Estimator (PE)}
The Progress Estimator provides a per-category estimate of \emph{``how much useful information remains now.''}
We instantiate PE in two interchangeable forms: (a) a heuristic expected discrete-gain variant that blends
informativeness rate, residual completeness deficit, and semantic ``saturation''; and (b) an information-theoretic
variant that replaces the deficit with binary entropy to yield expected information gain (bits) per category.

\subsubsection{Common Components}
Both variants share the following components.

\noindent \textbf{(1) Informativeness rate.}
We estimate the empirical probability that a query about category $i$ yields information gain using additive (Laplace) smoothing:
\begin{equation}\label{eq2:empirical_probability}
\rho_i(t)=\frac{k_i(t)+a}{m_i(t)+a+b},\qquad (a=b=1).
\end{equation}
Intuitively, $\rho_i(t)$ rises when queries to $i$ are frequently informative and falls when repeated queries fail to produce updates.

\noindent \textbf{(2) Semantic deficit.}
To model semantic saturation, we define a normalized semantic deficit:
\begin{equation}\label{eq3:semantic_deficit}
\psi_i(t)=1-\frac{\|e_i(t)\|}{\max_{j\in\mathcal{M}}\|e_j(t)\|+\varepsilon_e}\ \in[0,1],
\end{equation}
where $\varepsilon_e>0$ prevents division by zero early in a dialogue. Here $\psi_i(t)\approx 0$ when $i$ is relatively saturated
(high embedding trace magnitude) and $\psi_i(t)\approx 1$ when the semantic trace is comparatively sparse.

\noindent \textbf{(3) Importance and dependencies.}
Each category $i$ has a fixed importance weight $w_i\in[0,1]$, and a dependency gate $g_i(t)\in[0,1]$ that can encode prerequisites
(e.g., suppressing a category until required categories reach specified levels). PE multiplies by $w_i\,g_i(t)$.

\subsubsection{Heuristic $\mathrm{PE}$ (Expected Discrete Gain)}
The heuristic variant uses discrete completeness deficit. For each $i\in\mathcal{M}$:
\begin{equation}\label{eq4:heuristic_PE}
\begin{multlined}
\mathrm{PE}^{\text{H}}_{i}(t)=
\Big[\alpha\ \underbrace{\rho_i(t)\big(1-\upsilon_i(t)\big)}_{\substack{\text{expected discrete}\\\text{gain}}}
+(1-\alpha)\ \underbrace{\psi_i(t)}_{\substack{\text{semantic}\\\text{deficit}}}\Big]\cdot w_i\cdot g_i(t),
\end{multlined}
\end{equation}
where $\alpha\in[0,1]$ balances discrete and semantic signals. The term $\rho_i(t)\big(1-\upsilon_i(t)\big)$ is the expected discrete
gain. If the category is frequently informative ($\rho_i$ high) and incomplete ($1-\upsilon_i$ high), then residual potential is high.

\subsubsection{Formal $\mathrm{PE}$ (Shannon-based EIG)}
The formal variant replaces the discrete deficit with an information-theoretic uncertainty signal.
Let $L_i(t)\in\{\text{resolved},\text{unresolved}\}$ be a binary latent variable for category $i$ with belief
$p_i(t)=\Pr(L_i(t)=\text{resolved}\mid\text{context})\approx \upsilon_i(t)$. Uncertainty is quantified by binary entropy (bits):
\begin{equation}\label{eq5:Shannon_binary_entropy}
H_b\!\big(p_i(t)\big)=-p_i(t)\log_2 p_i(t)-\big(1-p_i(t)\big)\log_2\big(1-p_i(t)\big).
\end{equation}
High entropy indicates substantial uncertainty remains, while low entropy indicates little uncertainty left to remove.
We then compute expected information gain in bits:
\begin{equation}\label{eq6:formal_PE}
\begin{multlined}
\mathrm{PE}^{\text{E}}_{i}(t)=
\Big[\alpha\ \underbrace{\rho_i(t)\,H_b\!\big(p_i(t)\big)}_{\substack{\text{expected information}\\\text{gain (bits)}}}
+(1-\alpha)\ \underbrace{\psi_i(t)}_{\substack{\text{semantic}\\\text{deficit}}}\Big]\cdot w_i\cdot g_i(t).
\end{multlined}
\end{equation}
Here $\rho_i(t)H_b(p_i(t))$ is an outcome-aware expected information signal of uncertainty (bits) weighted by the empirical chance that
queries to $i$ tend to yield updates.

Both $\mathrm{PE}^{\text{H}}$ and $\mathrm{PE}^{\text{E}}$ share the same outer structure ($w_i\,g_i(t)$) and differ only in the
information signal inside the brackets. Discrete deficit $\rho_i(1-\upsilon_i)$ versus entropy $\rho_i H_b(p_i)$.
This symmetry supports clean ablations without altering the surrounding method.

\subsection{Stalling Index (SI): Interaction Health Telemetry}
\label{subsec:stall_index}

\subsubsection{Design Philosophy: Observable Failure Signatures}
Productive information-gathering dialogues exhibit two observable properties:
(1) queries yield responses with measurable information content, and
(2) successive queries explore semantically distinct aspects.
When these degrade (i.e., repeated category probing yields semantically similar responses with minimal knowledge gain) the interaction has entered an unproductive state we term \emph{conversational stalling}.

SI monitors this pattern through observable turn-level features. When a category is repeatedly queried within a short window with high semantic similarity between exchanges and minimal knowledge gain, SI rises. This design prioritizes actionable signals over causal attribution, recognizing that field systems may lack reliable methods for diagnosing whether unproductive patterns stem from questioning strategy, interviewee limitations, or other factors. By focusing on observable patterns rather than underlying causes, SI provides a pragmatic monitoring signal suitable for autonomous systems where causal diagnosis infrastructure (e.g., human affect recognition, cognitive load estimation) may be absent or unreliable.

\subsubsection{Intuition}
Re-asking the same category while receiving semantically similar answers and little discrete gain indicates unproductive exploration.
We detect stalling by fusing (i) discrete repetition cues in a short trailing window and (ii) embedding-space movement: high cosine
similarity between successive category embeddings indicates low semantic drift.

\subsubsection{Windowed sets and counts}
Let $\mathcal{W}(t)=\{t-W+1,\ldots,t\}$ denote the trailing window (for $t\ge W$).
Let $c_i(t)=\sum_{\tau\in\mathcal{W}(t)}\mathds{1}\{i_\tau=i\}$ be the number of times category $i$ was queried in the last $W$ turns.
Let $r_{\min}\in\mathbb{N}$ be the repetition threshold and define the set of repeated categories
$\mathcal{R}(t)=\{\, i\in\mathcal{M}\mid c_i(t)\ge r_{\min}\,\}$.
Let $i_{\text{recent}}:=i_t$ denote the category queried at the current turn.
For any $i$, let $\tau_i(t)=\max\{\tau\in\mathcal{W}(t)\mid i_\tau=i\}$ be the most recent time $i$ was queried in the window, and define
$\Delta\upsilon_i^{\text{recent}}(t)=\Delta\upsilon_i(\tau_i(t))$.

\subsubsection{Blended definition}
We combine discrete repetition patterns with semantic movement:
\begin{equation}\label{eq7:SI_blend}
\mathrm{SI}(t)=\beta\,\mathrm{SI}^{\text{disc}}(t)+(1-\beta)\,\mathrm{SI}^{\text{sem}}(t),\qquad \beta\in[0,1].
\end{equation}
Both components share a gain-dampening function
\[
\mathcal{D}(\Delta\upsilon;\lambda)=1-\min\{1,\lambda\,\Delta\upsilon\},
\]
which suppresses stall signals when recent knowledge gain is substantial. Here $\lambda>0$ controls dampening strength.

\noindent \textbf{Discrete component.}
Let $r_{\max}(t)=\max_{i\in\mathcal{M}} c_i(t)$ be the maximum repetition count in the window. We scale repetition frequency by gain dampening:
\begin{equation}\label{eq8:si_disc}
\mathrm{SI}^{\text{disc}}(t)=
\underbrace{\frac{\max\{0,\, r_{\max}(t)-1\}}{W}}_{\text{repetition fraction}}
\cdot
\underbrace{\mathcal{D}\big(\Delta \upsilon_{i_{\text{recent}}}^{\text{recent}}(t); \lambda\big)}_{\text{gain dampening}}.
\end{equation}
When no category repeats ($r_{\max}=1$), we have $\mathrm{SI}^{\text{disc}}=0$.

\noindent \textbf{Semantic component.}
For each repeated category $i\in\mathcal{R}(t)$, we assess semantic movement by comparing the current embedding with the previous time
category $i$ was queried. Let $e_i(t)$ be the current embedding trace and $e_i^{\text{prev}}$ be the trace stored at the previous query of $i$.
Cosine similarity near 1 indicates successive answers about $i$ are semantically similar:
\begin{equation}\label{eq9:cosine}
\cos\!\big(e_i(t),e_i^{\text{prev}}\big)=
\frac{\langle e_i(t),e_i^{\text{prev}}\rangle}{\|e_i(t)\|\,\|e_i^{\text{prev}}\|+\epsilon},
\end{equation}
where $\epsilon>0$ prevents division by zero. We map cosine similarity from $[-1,1]$ to $[0,1]$, then apply per-category gain dampening:
\begin{equation}\label{eq10:si_sem_each}
s_i^{\text{sem}}(t)=
\underbrace{\frac{1+\cos\!\big(e_i(t),e_i^{\text{prev}}\big)}{2}}_{\text{high if little semantic drift}}
\cdot
\underbrace{\mathcal{D}\big(\Delta \upsilon_{i}^{\text{recent}}(t); \lambda\big)}_{\text{gain dampening}}.
\end{equation}
The semantic component averages over repeated categories:
\begin{equation}\label{eq11:si_sem}
\mathrm{SI}^{\text{sem}}(t)=
\begin{cases}
\frac{1}{|\mathcal{R}(t)|}\sum_{i\in\mathcal{R}(t)} s_i^{\text{sem}}(t), & \text{if } \mathcal{R}(t)\neq\emptyset,\\
0, & \text{otherwise}.
\end{cases}
\end{equation}
By construction, $\mathrm{SI}^{\text{disc}},\mathrm{SI}^{\text{sem}}\in[0,1]$ and thus $\mathrm{SI}(t)\in[0,1]$.

\textit{Example.}
Consider a $W=3$ window where ``location'' is queried twice. If the second location query yields an embedding highly similar to the first
(e.g., cosine similarity $\simeq 0.9$) and produces minimal knowledge gain (e.g., $\Delta\upsilon\simeq 0.02$), then
(i) $\mathrm{SI}^{\text{disc}}$ reflects repetition with little gain, and (ii) $\mathrm{SI}^{\text{sem}}$ reflects high semantic similarity.
Both contribute to a higher $\mathrm{SI}(t)$, signaling potential stalling.

\subsubsection{Implementation details}
We use $W=3$, repetition threshold $r_{\min}=2$, SI detection threshold $\theta=0.20$,
$\alpha=0.5$, and $\lambda=5.0$.
We set the blend parameter $\beta=0.4$ for synthetic embeddings and $\beta=0.5$ for transformer-based embeddings.
Additional mechanisms include category-dependent scaling factors for transformer models and compression for values above $0.8$ for numerical stability.

\begin{algorithm}[htbp]
\caption{Dialogue Telemetry ($\mathrm{DT}$) Monitoring}
\label{alg1:monitoring}
\begin{algorithmic}[1]
\REQUIRE Dialogue sequence $D = \{(q_1,y_1), \ldots, (q_T,y_T)\}$
\ENSURE Turn-level PE and SI telemetry traces
\STATE Initialize hybrid state $s(0)$ (Eq.~\ref{eq1:hybrid_state})
\STATE Set window $W$, repetition threshold $r_{\min}$, SI detection threshold $\theta$
\FOR{each turn $t = 1$ to $T$}
    \STATE Process question--answer pair $(q_t, y_t)$ and identify targeted category $i_t$
    \STATE Update hybrid state $s(t)$ from $y_t$ (counts, completeness gains, embeddings)
    \FOR{each category $i \in \mathcal{M}$}
        \STATE Compute $\mathrm{PE}_i^{\bullet}(t)$ (Eq.~\ref{eq4:heuristic_PE} or Eq.~\ref{eq6:formal_PE})
    \ENDFOR
    \STATE Compute $\mathrm{SI}(t)$ (Eq.~\ref{eq7:SI_blend})
    \IF{$t \geq W$ and $\mathrm{SI}(t) > \theta$}
        \STATE Flag conversational stalling within window $\mathcal{W}(t)$
    \ENDIF
    \STATE Log telemetry metrics
\ENDFOR
\RETURN telemetry traces and detected stall windows
\end{algorithmic}
\end{algorithm}

\subsection{DT as Supervisory Controller Interface}\label{sec:supervisory-controller}
DT provides observables that enable closed-loop dialogue policies. While question realization (template vs.\ LLM generation)
and domain-specific strategy libraries are actuator-layer concerns outside DT's scope, the telemetry signals support generic
supervisory interventions such as:
\begin{itemize}[noitemsep,topsep=0pt]
\item \textbf{Category switching:} redirect to categories with higher residual potential; abandon a category when $\mathrm{SI}(t)>\theta$.
\item \textbf{Strategy adaptation:} switch interviewing strategy when $\mathrm{SI}(t)$ indicates revisitation.
\item \textbf{Human operator escalation:} sustained high $\mathrm{SI}(t)$ with elevated PE triggers handoff to a human operator or dashboard intervention.
\item \textbf{Modality switching:} high $\mathrm{SI}(t)$ can trigger transitions to maps/photos/sensor-based evidence when available.
\item \textbf{Termination:} stop when all $\upsilon_i$ exceed resolution thresholds, or when residual potential is low with $\mathrm{SI}(t)$ remaining low;
treat sustained $\mathrm{SI}(t)>\theta$ as a breakdown signal requiring strategy change or handoff.
\end{itemize}
These interventions leverage DT's model-agnostic design. In other words, the same telemetry can drive control regardless of whether the underlying
question generator uses templates, retrieval-augmented generation, or end-to-end neural policies.
Our RL integration (Sec.~\ref{sec:exp2}) demonstrates one instantiation of this interface contract, but DT could support diverse control architectures.

\begin{figure*}[t]
\centering
\includegraphics[width=\textwidth]{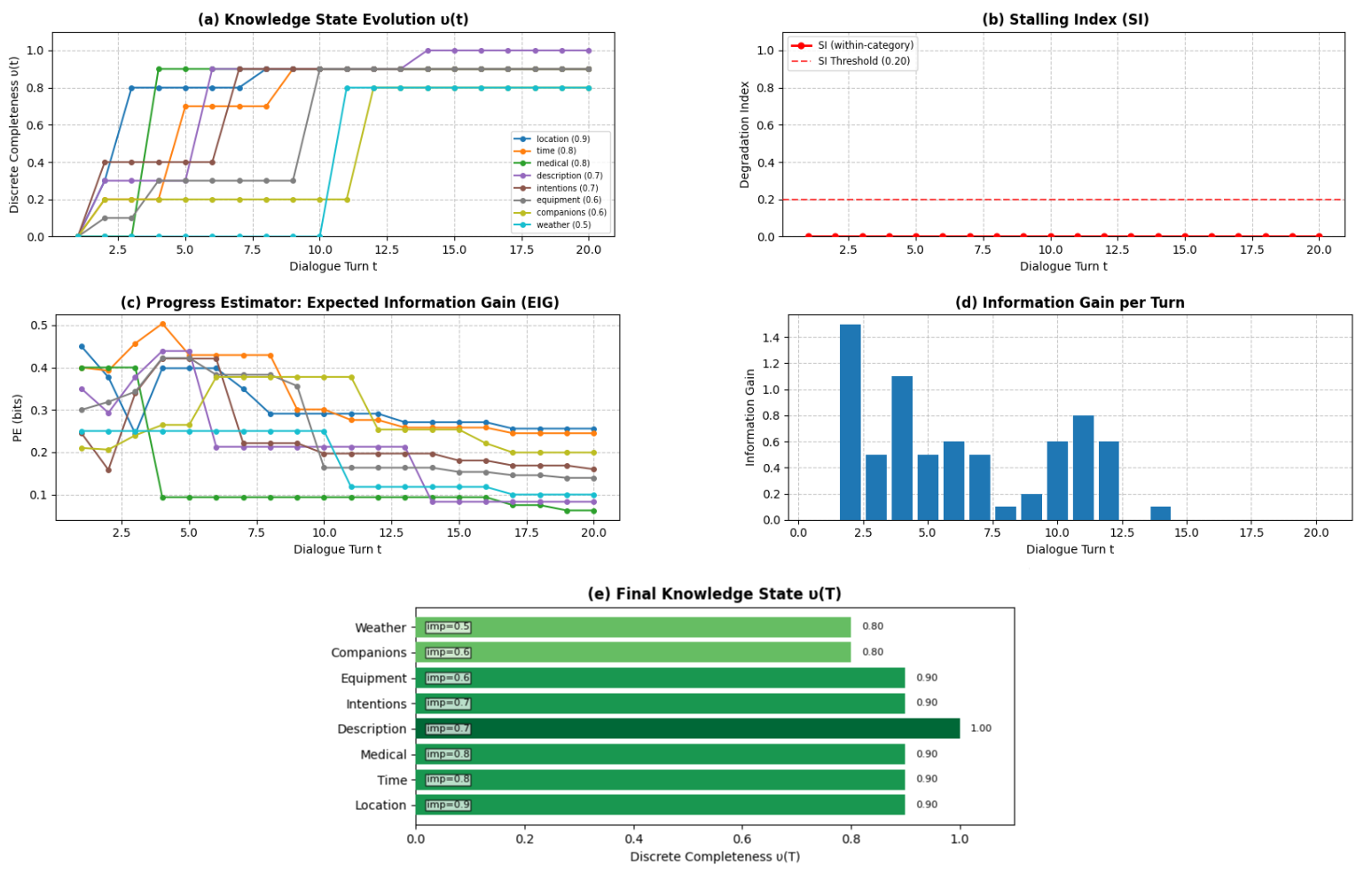}
\caption{DT Telemetry Monitoring of Efficient Dialogue. Turn-by-turn tracking of optimal 20-turn \(\mathrm{DT}\)-guided interrogation. (a) Systematic knowledge evolution achieving high completion rates; (b) Stall Index (\(\mathrm{SI}\)) remaining below detection threshold throughout; (c) EIG values decreasing systematically as gaps close; (d) Information gain following diminishing returns pattern; (e) Final state showing successful completion aligned with importance weights.}
\label{fig2:results_4-2-efficient}
\end{figure*}

\begin{algorithm}[htbp]
\caption{RL with Dialogue Telemetry ($\mathrm{DT}$) signals}
\label{alg2:rl_integration}
\begin{algorithmic}[1]
\REQUIRE Training episodes $N$, maximum turns per episode $T$
\ENSURE Trained dialogue policy $\pi$
\STATE Initialize PPO agent with policy $\pi$ and value function $V$
\FOR{episode $n = 1$ to $N$}
    \STATE Initialize hybrid state $s(0)$ with categories $\mathcal{M}$ (Eq.~\ref{eq1:hybrid_state})
    \STATE $t \leftarrow 0$, terminated $\leftarrow$ False
    \WHILE{$t < T$ and not terminated}
        \STATE Compute $\mathrm{PE}_i^{\bullet}(t)$ for all $i \in \mathcal{M}$ (Eq.~\ref{eq4:heuristic_PE} or Eq.~\ref{eq6:formal_PE})
        \STATE Compute $\mathrm{SI}(t)$ (Eq.~\ref{eq7:SI_blend})
        \STATE Build observation $z(t)$ from $s(t)$ (Eq.~\ref{eq14:augmented_observation})
        \STATE Select action $a_t \sim \pi(\cdot \mid z(t))$ \COMMENT{$a_t$ is a strategy--category pair}
        \STATE Realize question $q_t \leftarrow \textsc{Realize}(a_t)$ \COMMENT{e.g., template realization}
        \STATE Environment returns answer $y_t$; update hybrid state $s(t{+}1)$
        \STATE Compute reward $R(t)$ (Eq.~\ref{eq15:reward_shaping})
        \IF{termination criterion (Eq.~\ref{eq12:termination}) is satisfied}
            \STATE terminated $\leftarrow$ True
        \ENDIF
        \STATE $t \leftarrow t + 1$
    \ENDWHILE
    \STATE Update policy $\pi$ and value function $V$ using PPO
\ENDFOR
\RETURN trained policy $\pi$
\end{algorithmic}
\end{algorithm}

\section{Experimental Studies}
We evaluate \(\mathrm{DT}\) through two complementary experimental paradigms using a shared SAR witness interview simulation. Both experiments operate within the same domain configuration and dialogue environment, but serve different evaluation objectives: standalone conversation monitoring validates \(\mathrm{DT}\)'s analysis capabilities, while RL integration demonstrates \(\mathrm{DT}\)'s effectiveness as guidance signals for learning question policies. We note that throughout our experimental studies, we use the formal \(\mathrm{PE}\).

\subsection{Experimental Setup}

\textbf{Domain \& categories.} Both studies use a simulated SAR witness-interview setting in which time-critical information gathering achieved across eight predefined knowledge categories: location, time, description, equipment, medical, weather, companions, and intentions. Each category has an operational weight \(w_i\) reflecting field criticality \cite{9980686} (location (0.9), time (0.8), medical (0.8), description (0.7), intentions (0.7), equipment (0.6), companions (0.6), weather (0.5)).


\textbf{Question Strategies.} We employ six interviewing strategies adapted from cognitive-interview practice such as Free Recall, Specific Detail, Clarification, Context Reinstatement, Summary Confirmation, and Rapport Building. Each strategy is instantiated through pre-defined question templates that produce strategy \(\times\) category actions consistent with interviewing best practices.


\textbf{Pre-generated dialogue corpus.} We construct a complete dialogue corpus offline using LLM-generated content to ensure realistic conversational variability. We used \texttt{gemma3:27b} \cite{gemmateam2025gemma3technicalreport}. Question templates for each strategy–category pair were generated using targeted prompts, and witness responses were synthesized with varied information density. This pre-generated corpus contains 25\(+\) curated witness responses that mimic the intertwined, cross-category evidence patterns found in real SAR interviews. All content generation occurs during setup, not at runtime.

\textbf{Knowledge scoring methodology.} Each answer–category pair in our pre-generated corpus receives a knowledge contribution score \(\upsilon\in[0,1]\) based on completeness and specificity. Manual annotations serve as primary labels, validated by an LLM-based scorer using structured evaluation rules.

\textbf{Simulation Environment.} The shared experimental environment operates deterministically over our pre-generated corpus, providing witness responses that span low to high information yield (knowledge gains typically ranging from $\sim$0.01 to 0.8\(+\)). The simulator maintains (i) discrete knowledge accumulation per category (the \(\upsilon_i\) trace) and (ii) semantic embeddings for each response, enabling both uncertainty calculations and semantic revisitation analysis.

\textbf{Semantic Embeddings.} All sentence embeddings for corpus responses are precomputed using the \texttt{all-MiniLM-L6-v2} \cite{wang2020minilmdeepselfattentiondistillation} and normalized to unit length. This precomputation approach stabilizes similarity calculations and ensures faster and consistent SI calculations across experiments.

\textbf{Reproducibility notes.} Prompts for question generation and scoring, along with implementation details for the simulation and telemetry, are available as supplementary materials; Sections~\ref{sec:exp1} and \ref{sec:exp2} specify experiment-specific settings on top of this shared setup.

\begin{figure*}[t]
\centering
\includegraphics[width=\textwidth]{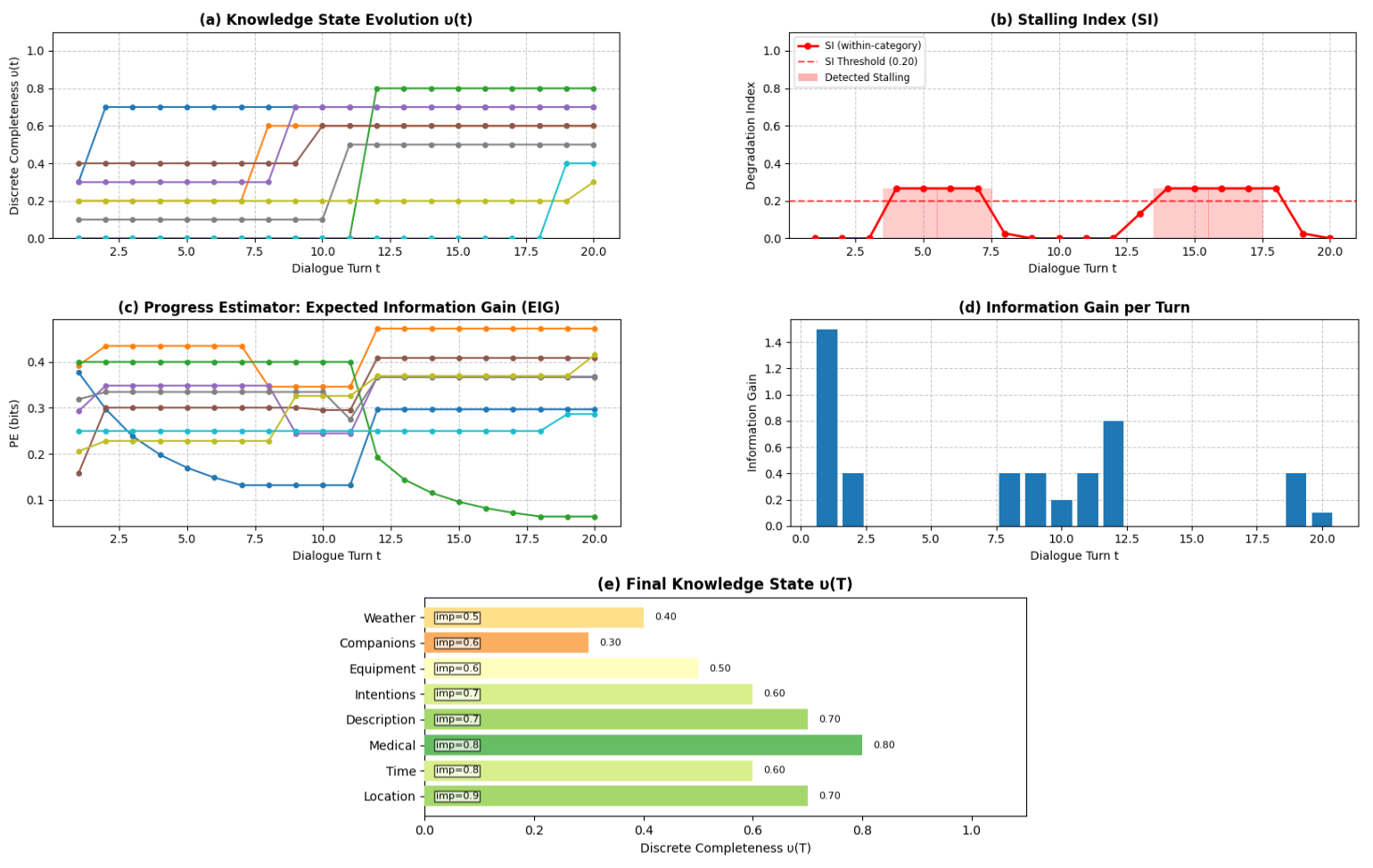}
\caption{DT Telemetry Detection of Conversational Stalling. Monitoring of a 20-turn dialogue exhibiting dual stalling incidents. (a) Knowledge evolution showing suboptimal progression; (b) \(\mathrm{SI}\) detecting two vortex periods (red shaded) exceeding threshold; (c) EIG values remaining persistently high due to unresolved information gaps; (d) Information gain showing near-zero acquisition during stalling periods; (e) Final state revealing poor completion across categories.}
\label{fig3:dual}
\end{figure*}

\subsection{Experiment 1: Conversation Monitoring}
\label{sec:exp1}
The first experiment evaluates \(\mathrm{DT}\)'s capability to monitor and analyze dialogue transcripts, independent of any learning system. Using ground-truth labeled dialogue sequences from the shared SAR simulation environment, we test the method's accuracy in identifying information-gathering progress and detecting conversational inefficiencies.

\subsubsection{\underline{Objective}}
We test \(\mathrm{DT}\)'s ability to: (1) quantify information-gathering efficiency through turn-level EIG measurements, (2) detect conversational stalling incidents (we compare against known ground-truth stalling periods), and (3) distinguish between productive and unproductive dialogue segments with measurable accuracy.

\subsubsection{\underline{Monitoring Implementation}} 
We implement standalone monitoring using Algorithm~\ref{alg1:monitoring}. The system processes dialogue turns sequentially, computing PE and SI values after each exchange and detecting conversational stalling behavior when SI exceeds 0.20. We construct two controlled dialogue scenarios by sampling question--answer combinations with known ground-truth stalling labels: 
1) \textit{Efficient Dialogue (Scenario 1):} A 20-turn conversation following efficient information-gathering patterns. Ground truth: zero stalling episodes. 
2) \textit{Stalling Dialogue (Scenario 2):} A 20-turn conversation with two deliberate within-category stalling episodes where the agent repeatedly probes the same categories (turns 5--7 for location; turns 13--17 for medical), yielding minimal knowledge gains ($\leq 0.05$ per turn). 
Ground-truth stalling periods are defined as consecutive turns targeting the same category with knowledge gains $\leq 0.05$ per turn. We measure information efficiency as cumulative knowledge gain per turn, stall detection accuracy using the SI threshold, knowledge completion rates across categories, and detection timing for conversational cycling.

\subsubsection{\underline{Results: Interaction Health Detection}}

We validate SI's detection capability through two controlled scenarios:

\paragraph{Scenario 1: Efficient dialogue baseline.}
Figure~\ref{fig2:results_4-2-efficient} shows DT telemetry in an efficient dialogue. 
Knowledge-state evolution exhibits systematic progression across high-importance categories, 
PE/EIG decreases monotonically as gaps close, and SI remains below the 0.20 detection threshold 
throughout (panel b). Final knowledge state (panel e) achieves high completion rates across all 
dimensions (0.75--0.90), validating that DT produces no false positives during productive 
information gathering.

\paragraph{Scenario 2: Stalling dialogue.}
Figure~\ref{fig3:dual} shows DT telemetry detecting unproductive interaction patterns. SI 
identifies two distinct stalling episodes (turns 5--7 and 13--17, panel b, red shaded regions) 
where repeated category probing yields semantically similar, low-marginal-gain responses. During 
these episodes, knowledge gains decline from initial values of 0.70--0.80 to near-zero (0.0--0.02). 
Correspondingly, PE/EIG remains elevated for unresolved categories (panel c), reflecting persistent 
residual information potential that is not being efficiently harvested. Final knowledge state 
(panel e) shows suboptimal completion: weather (0.40), companions (0.30), equipment (0.50), and 
intentions (0.60) remain underexplored.

\paragraph{Detection summary.}
SI successfully detects both injected stalling episodes (2/2) and produces no false positives in the 
efficient baseline (0/20 turns flagged). Average SI values during stalling episodes range from 
0.25--0.45 (above 0.20 threshold), while remaining below 0.05 during efficient dialogue. These 
controlled results demonstrate that SI reliably flags interaction patterns characterized by 
repeated probing with diminishing returns.

\begin{figure*}[!t]
    \centering
    
    \begin{subfigure}{0.45\textwidth}
        \centering
        \includegraphics[width=\linewidth]{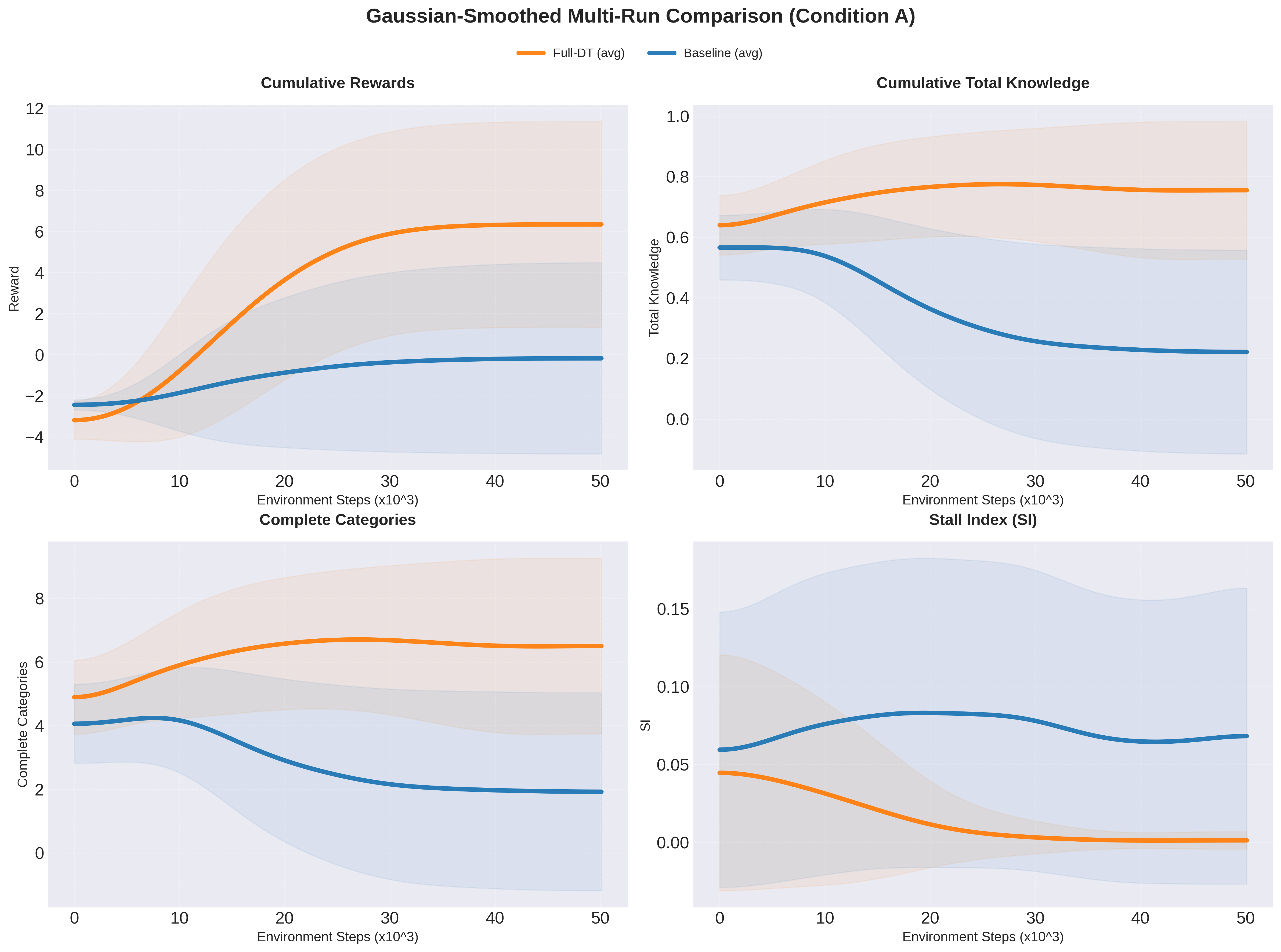}
        \caption{RL Integration - Condition A.}
        \label{fig:RL-terminationA}
    \end{subfigure}
    \hspace{0.05\textwidth}  
    \begin{subfigure}{0.45\textwidth}
        \centering
        \includegraphics[width=\linewidth]{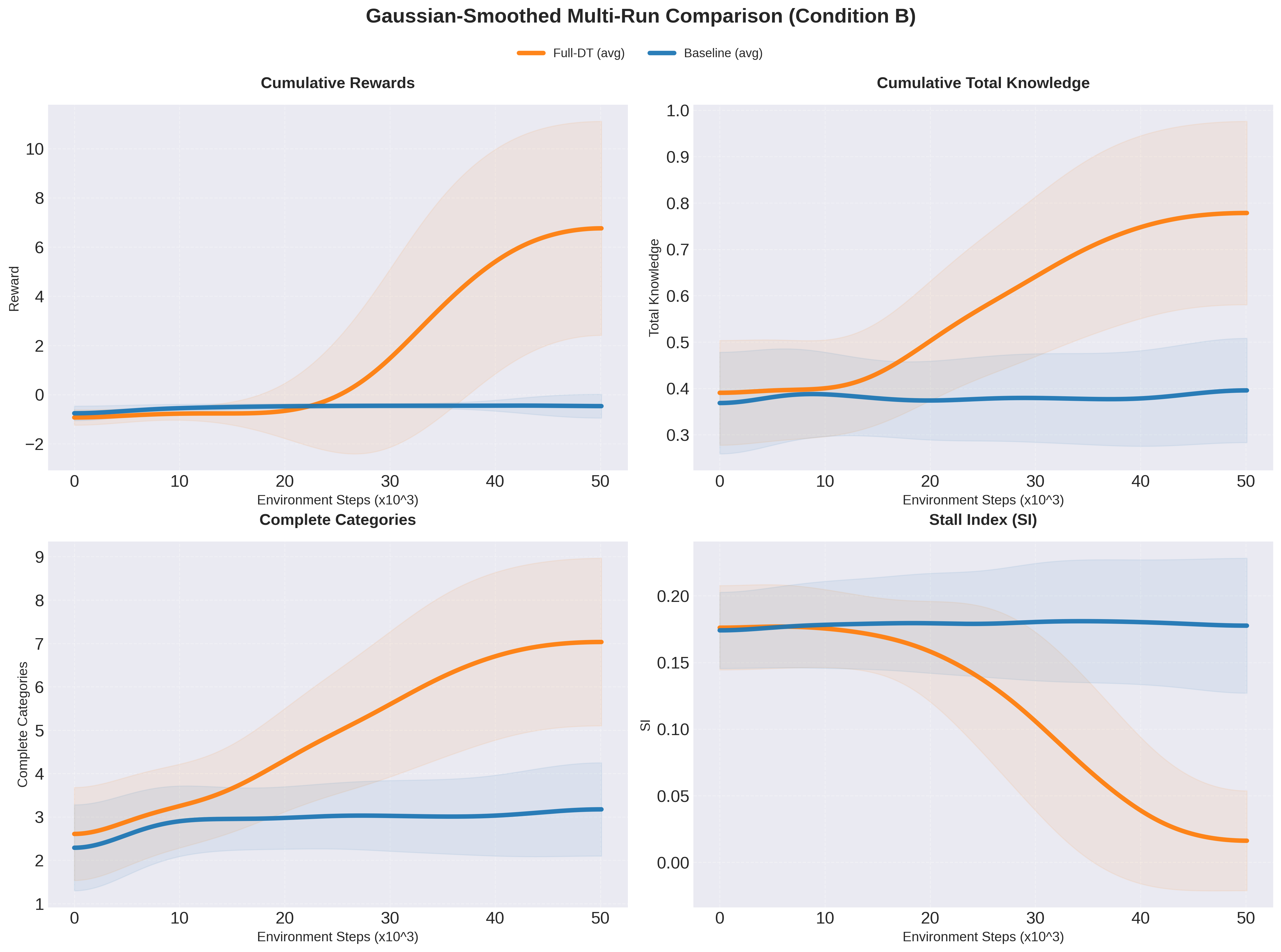}
        \caption{RL Integration - Condition B.}
        \label{fig:RL-terminationB}
    \end{subfigure}
    
    \vspace{0.5em}
    
    \begin{subfigure}{0.45\textwidth}
        \centering
        \includegraphics[width=\linewidth]{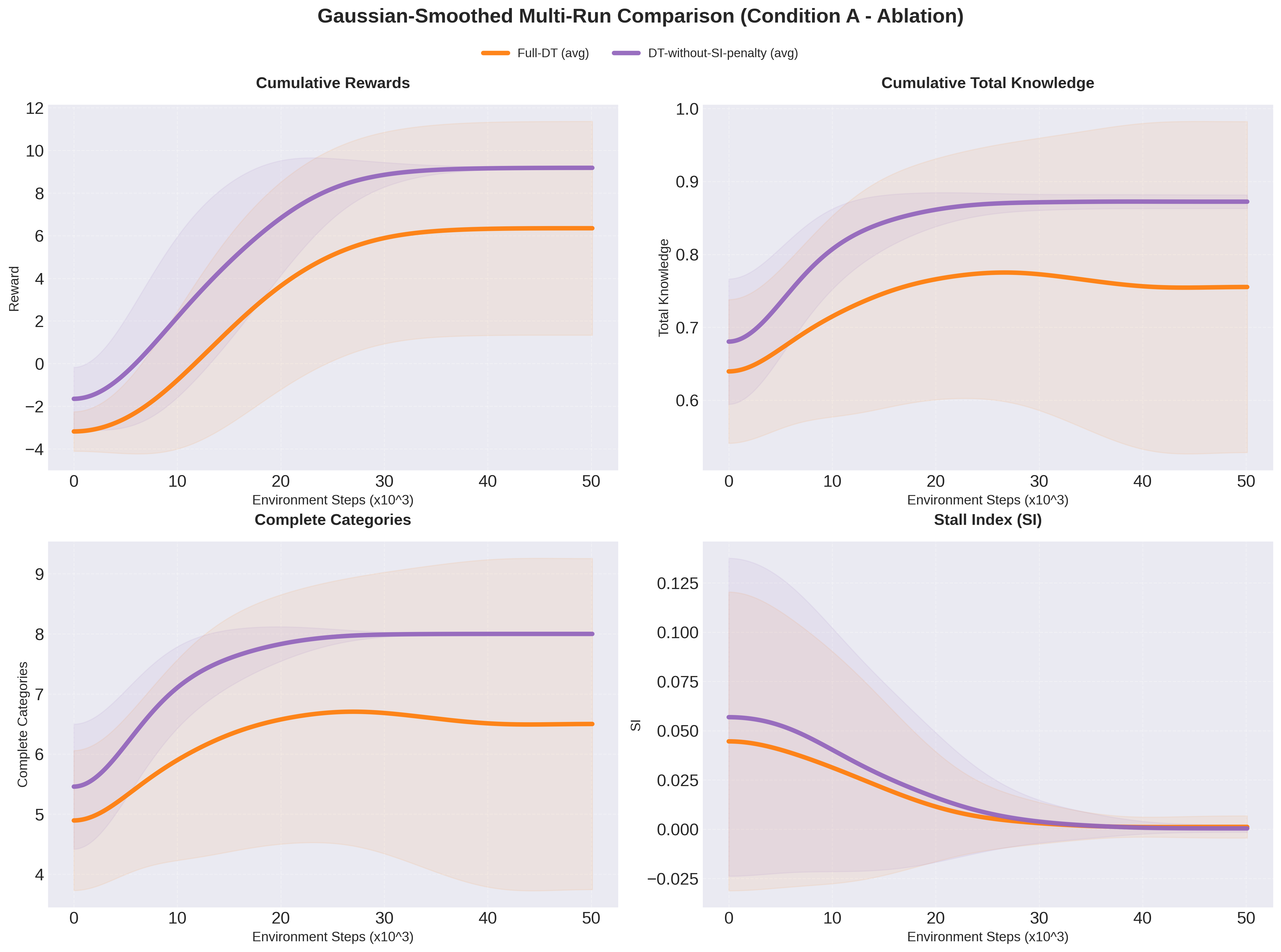}
        \caption{Condition A Ablation.}
        \label{fig:Ablation_conditionA}
    \end{subfigure}
    \hspace{0.05\textwidth}  
    \begin{subfigure}{0.45\textwidth}
        \centering
        \includegraphics[width=\linewidth]{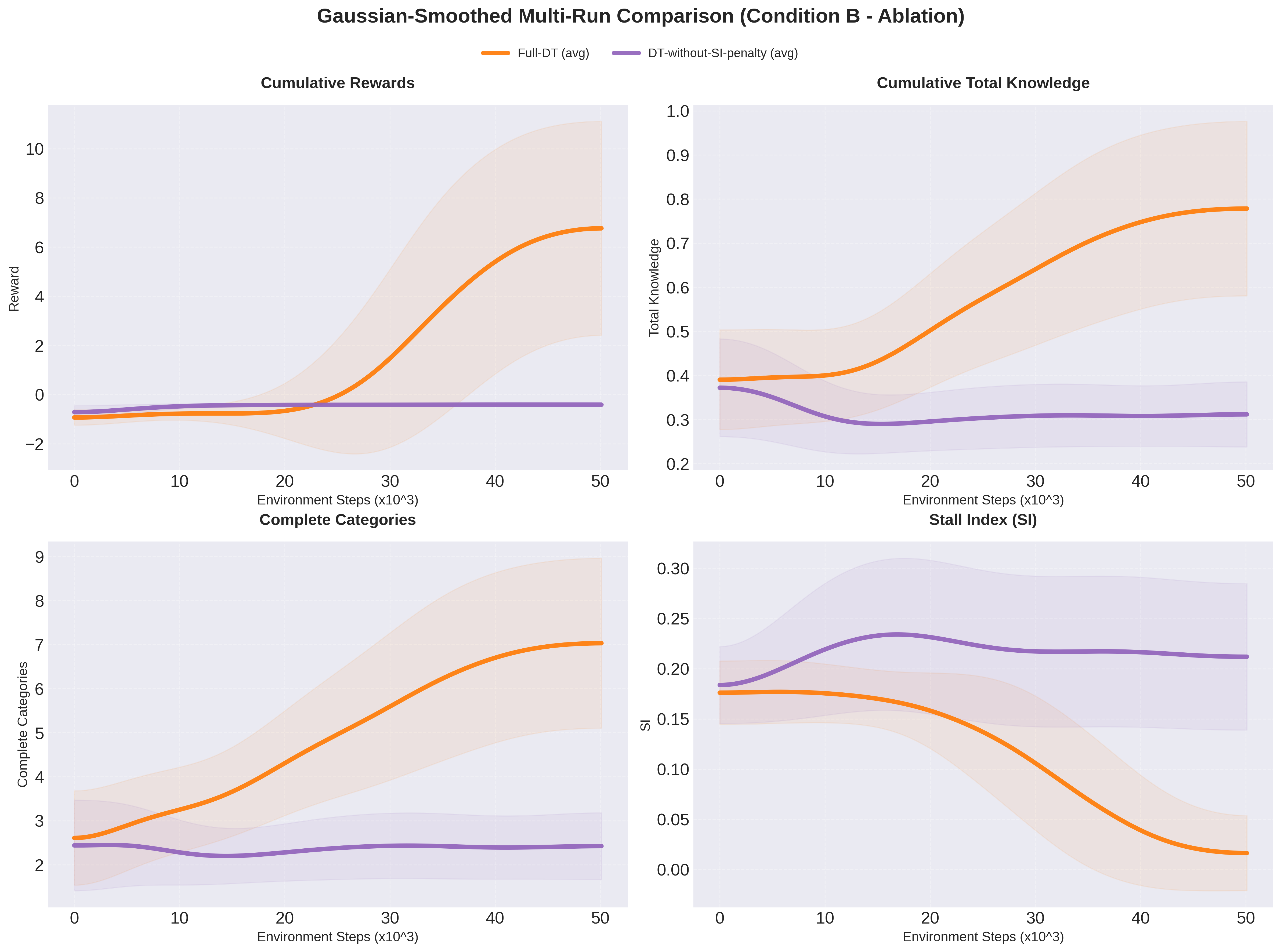}
        \caption{Condition B Ablation.}
        \label{fig:Ablation_conditionB}
    \end{subfigure}
    
    \caption{RL Integration and Ablation results under Conditions A and B. (a) Condition A (standard termination): Full-DT vs Baseline. (b) Condition B (stall-aware termination): Full-DT vs Baseline. (c) Condition A ablation: Full-DT vs DT-without-SI-penalty. (d) Condition B ablation: Full-DT vs DT-without-SI-penalty. Four metrics track agent performance over 50k timesteps: Episode Rewards (higher = better, indicating successful task completion), Total Knowledge (higher = better, measuring information acquisition efficiency), Complete Categories (higher = better, showing number of resolved knowledge categories out of 8 total), and Stall Index/SI (lower = better, indicating reduced conversational repetition). All curves represent Gaussian-smoothed averages across 25 independent training runs with \(\pm 1\sigma\) confidence bands.}
    \label{fig:RL-ablation-all}
\end{figure*}

\subsection{Experiment 2: RL Integration}
\label{sec:exp2}
Having validated that DT signals accurately track dialogue health in fixed traces, we next evaluate if these signals can actively guide an agent to learn better policies. The second experiment operationalizes the \(\mathrm{DT}\) inside an episodic RL environment and evaluates its effectiveness as guidance signals for automated dialogue policies within the same SAR simulation environment. Here, RL serves as a case study to illustrate \(\mathrm{DT}\)'s model-agnostic utility in influencing behavior toward fewer stalls and higher efficiency, particularly when repetition risks dialogue breakdown. Our claim is that telemetry's value emerges specifically when stall incurs costs, not from RL architectural advantages.

\subsubsection{\underline{RL Scenario}}
We model the decision process as a partially observable Markov decision process (POMDP) where the agent selects actions as strategy-category pair (47 actions) to elicit answers, update knowledge, and progress toward schema completion. Answers trigger discrete knowledge gains $\Delta\upsilon_i$ and semantic embedding updates, advancing the hybrid state $s$ (Eq.~(\ref{eq1:hybrid_state})). The base task reward encourages efficient completion while penalizing steps, with episodes limited to \(T\)=25 turns or until all categories are resolved,

\begin{equation}\label{eq12:termination}
    \forall i\in \mathcal{M}: \quad \upsilon_i(t) \geq\tau_i \qquad \text{with}\ \tau_i \in (0,1) \ (\text{default 0.8)}
\end{equation}

\subsubsection{\underline{Agent Variants}}
To isolate telemetry's utility, we compare two agent configurations that differ only in observation space and reward structure:

\textbf{Baseline (no telemetry).} The baseline agent operates without telemetry awareness, receiving minimal observations

\begin{equation}\label{eq13:minimal_observation}
z(t)
\;=\;
\Big[
\upsilon_{1:|\mathcal{M}|},\ t/T
\Big]
\in [0,1]^{\,|\mathcal{M}|+1}
\end{equation}

\textbf{Full-DT (telemetry-guided).} The telemetry agent receives full \(\mathrm{DT}\) signals as augmented observations.

\begin{equation}\label{eq14:augmented_observation}
z(t)
\;=\;
\Big[
\underbrace{p_{1:|\mathcal{M}|}}_{\text{beliefs }p_i\approx \upsilon_i},\ 
\underbrace{\mathrm{PE}^{\bullet}_{1:|\mathcal{M}|},\ \mathrm{SI}}_{\text{telemetry}},\ t/T
\Big]
\in [0,1]^{\,2|\mathcal{M}|+2}
\end{equation}
In our setup $|\mathcal{M}|=8$, giving 9 and 18 dimensions for baseline and telemetry agents, respectively.

The base task reward $R_{\text{task}}(t)$ used by the baseline is augmented with a stall-aware shaping term for the Full-DT agent:
\begin{equation}\label{eq15:reward_shaping}
R_t \;=\; R_{\text{task}}(t)\;-\;\kappa\,\mathrm{SI}(t),\qquad \kappa\ge 0,
\end{equation}
where $\mathrm{SI}$ is computed from $s(t)$ as in Sec.~\ref{subsec:stall_index} and \(R_{\text{task}}=-c_\text{step} +b_\text{term}10\{\text{all} \ p_i(t) \geq \tau\}\) with default step penalty and terminal bonus +10 when the knowledge categories reach the resolution threshold; \(\kappa\) controls stall penalty strength. The training objective is the standard discounted return $\max_\theta \mathbb{E}\big[\sum_t \gamma^t R_t\big]$.

\subsubsection{\underline{Two-Condition Experimental Design}}
We evaluate both agents under two termination regimes that test different aspects of conversational dynamics: 

\textit{Condition A (Standard Termination).} Episodes follow Eq.~\ref{eq12:termination} and terminate when: (i) all categories reach resolution threshold  \(\upsilon_i\ge\) 0.8), or (ii) the turn budget is exhausted (\(T\)=25 turns). In this regime, stalling incurs efficiency costs but does not prevent knowledge acquisition. Agents can maximize knowledge values directly, making stall detection an additional feature rather than essential. \textit{Hypothesis A:} Under standard termination, telemetry agents should outperform baseline agents primarily through better exploration.

\textit{Condition B (Stall-Aware Termination).} Episodes terminate under Condition A rules plus: (iii) stall-triggered termination when \(\mathrm{SI}\geq0.10\) representing scenarios where stalling genuinely blocks success. This models witness frustration dynamics where repetitive questioning may lead to interview breakdown. \textit{Hypothesis B:} Under stall-aware termination, telemetry agents should significantly outperform baseline agents by learning to avoid conversational traps that baseline agents cannot detect.

\subsubsection{\underline{Implementation Details}}
Custom environment using OpenAI Gym interfaces. Each agent variant trains for 50,000 timesteps using PPO with standard hyperparameters. All experiments use 25 independent training runs per condition to ensure statistical reliability.

\subsubsection{\underline{Results}}

\textit{Condition A (Standard Termination):} Supporting Hypothesis A, Full-DT agent outperformed Baseline across all metrics (rewards, total knowledge, complete categories, stall progress) even without stall termination pressure (see Fig.~\ref{fig:RL-terminationA}). This demonstrates that DT actionable guidance provides systematic benefits beyond pure stall avoidance. The combined telemetry signals enable more efficient category completion despite increased complexity (i.e., larger observation space increases learning complexity). \textit{Condition B (Stall-Aware Termination):} Results strongly support Hypothesis B (see Fig.~\ref{fig:RL-terminationB}). Full-DT agents demonstrate clear monotonic learning across all metrics, successfully avoiding stalling patterns while maintaining productive information gathering. Baseline performance remains flat throughout training (essentially no learning occurs). The agent cannot detect when questioning patterns trigger stall termination, creating a failure mode where episodes end prematurely without meaningful progress signals. 


\subsubsection{\underline{Component Ablation Analysis}}
To isolate DT component contributions, we evaluate DT-without-SI-penalty (SI observation space but not in reward function).

\textit{Condition A -- Ablation.} DT-without-SI-penalty achieved optimal performance, surpassing Full-DT across all metrics (see Fig.~\ref{fig:Ablation_conditionA}). This reveals that SI penalties become counterproductive when stalling lacks terminal consequences, inducing over-conservative exploration that limits knowledge acquisition potential.

\textit{Condition B -- Ablation.} DT-without-SI-penalty failed to solve the task, with flat performance curves resembling baseline failure modes (see Fig.~\ref{fig:Ablation_conditionB}). Without explicit stall penalties, agents cannot learn to avoid the termination condition despite observing SI information. This confirms that explicit stall penalties are essential when stalling triggers episode termination. Comparative results averaged across 25 runs are shown in Tables~\ref{tab1:training_comparison_across} and~\ref{tab2:training_comparison_end}.

\section{Discussion}

\subsection{Takeaways from studies}
\textbf{Turn-level monitoring (Exp.~1).} DT tracked efficient interviews with steadily increasing completeness, monotonic EIG decay, and SI consistently below threshold (Fig.~\ref{fig2:results_4-2-efficient}). In contrast, the stalling dialogue showed two clear stalling periods where SI crossed the threshold, EIG stayed elevated (unresolved uncertainty), and per-turn gains collapsed, leading to substantially lower final completeness across several categories (Fig.~\ref{fig3:dual}). Together, these traces show that DT's signals are both diagnostic (flagging stalling when it starts) and explanatory (highlighting stalled progress as the combination of high residual uncertainty, low marginal gain, and semantic revisitation). SI successfully detected both stalling episodes (2/2) while producing no false positives during efficient dialogue (0/20 turns), demonstrating reliable pattern recognition in controlled settings.

\textbf{RL integration (Exp.~2).} With standard termination (Condition A), the Full-DT agent outperformed the Baseline across reward, total knowledge, and number of fully resolved categories despite the larger observation space. This indicates that PE and SI provide useful guidance during exploration beyond stall prevention (Fig.~\ref{fig:RL-terminationA}). Under stall-aware termination (Condition B), the Baseline essentially failed to learn while Full-DT learned steadily, confirming that explicit stall awareness is necessary when repetition carries operational costs (Fig.~\ref{fig:RL-terminationB}). The ablation isolates when to penalize SI. That said removing the penalty helps under Condition A but hurts under Condition B, where the penalty is required to avoid early termination (Figs.~\ref{fig:Ablation_conditionA} and~\ref{fig:Ablation_conditionB}).

These results suggest that systematic turn-level monitoring can help operationalize progress tracking and stall detection in time-sensitive domains, within the controlled SAR-inspired settings studied here. The ability to quantify ``bits removed per turn'' and detect conversational cycling provides interpretable progress indicators that could complement existing checklists and training protocols. DT's model-agnostic design also means the telemetry can augment existing systems without requiring architectural changes, supporting both automated policies and human-in-the-loop monitoring. The contextual utility findings from the ablation have immediate practical implications. Organizations can deploy observation-only SI for efficiency monitoring in low-stakes scenarios, while enabling stall penalties when repetition can block task completion.

\subsection{Instrumentation for Autonomous Information Gathering}
DT addresses a specific instrumentation gap in which autonomous systems conducting schema-grounded interviews lack the turn-level observables available in other control domains. PE quantifies residual information potential (where to probe next), while SI flags throughput degradation (when a line of questioning becomes unproductive). This supports:

\textbf{Closed-loop information acquisition.} Just as robots use joint encoders for manipulation and LiDAR for navigation, they can use PE and SI as dialogue telemetry. The RL integration (Sec.~\ref{sec:exp2}) demonstrates one control architecture that consumes these signals, but DT's model-agnostic design could support diverse implementations.

\textbf{Observable patterns vs.\ underlying causes.} SI flags throughput degradation through turn-level features such as repetition frequency, semantic similarity, and knowledge gain. It does not estimate \emph{why} degradation occurs (e.g., ineffective questioning, fatigue, distraction, or environmental factors). This scoping reflects a measurement philosophy under which the instrumentation should provide reliable signals under diverse conditions without requiring rich causal models. Just as a tachometer signals excessive RPM without diagnosing whether the cause is driver error or mechanical issues, SI signals throughput problems without attributing causation. Downstream controllers can then trigger appropriate interventions (strategy change, category switch, operator handoff) based on context and available information. This design choice prioritizes robustness since the same telemetry signals support monitoring regardless of whether inefficiency stems from questioning strategy, human state, environmental conditions, or their interaction.

\textbf{Compatibility with existing architectures.} DT operates on dialogue traces plus a task schema, making it compatible with template-based, retrieval-augmented, or neural question generators. The telemetry signals provide a controller-agnostic interface for supervisory interventions (Sec.~\ref{sec:supervisory-controller}).

\textbf{Deployment contexts.} By converting conversational dynamics into observable metrics, DT provides a layer of safety for deployments where human-level intuition is unavailable. We envisage two critical deployment contexts: (1) \emph{Autonomous First Contact} in disaster scenarios where autonomous agents may locate victims before human rescuers arrive. In this time-critical window, the system must independently guarantee that it is gathering actionable intelligence (high PE) without subjecting a traumatized victim to repetitive loops (high SI). DT provides the necessary signal to govern this interaction autonomously. (2) \emph{Tele-presence and Supervisory Control} for human-in-the-loop settings, where DT functions as a real-time efficiency dashboard. It allows a remote operator to monitor multiple autonomous interviewers simultaneously, flagging only those interactions that exhibit high stalling indices for human intervention.

The information-theoretic PE variant (bits) further enables unified planning across modalities. Comparing verbal-question EIG against sensor-based information gain can support multi-modal resource allocation in field robotics.

\subsection{Technical Insights}
\textbf{PE (bits) as a progress compass.} The entropy-based PE quantifies ``removable uncertainty now.'' In efficient dialogues it decays as gaps close. Conversely, when dialogues stall, PE remains high, signaling unresolved opportunities that the current line of questioning is not exploiting (Figs.~\ref{fig2:results_4-2-efficient},~\ref{fig3:dual}). In RL, policies learn to trade immediate PE against crossing category thresholds to reduce future turn consumption, explaining improved completion (Fig.~\ref{fig:RL-ablation-all}). PE serves as guidance rather than objective: while it estimates ``bits removable now,'' the learned value function can prefer longer-horizon actions that push categories past resolution thresholds over maximizing immediate gain. The augmented observation space resolving binary entropy's symmetry around \(p=0.5\) enables policies to distinguish \textit{probably incomplete} from \textit{probably complete} categories, supporting more strategic exploration.

\textbf{SI as early warning for throughput degradation.} SI rises when categories are repeatedly probed with minimal discrete gain and high embedding similarity. In transcripts, spikes align with low information gain and flat completeness curves. In RL, an SI-aware reward reshapes behavior away from locally repetitive probing toward productive category switches, especially when stalls cause failure (Condition B). The detection mechanism operates on observable turn-level patterns rather than internal state models, making it robust to diverse sources of inefficiency.

\subsection{Telemetry and Actuation Layers}
DT is an instrumentation layer, meaning it defines turn-level observables (PE, SI) and an interface contract that a dialogue policy can use for closed-loop decisions while remaining agnostic to how questions are realized. In Exp.~2 we intentionally instantiate the actuation layer using a small discrete move set (strategy\(\times\)category actions realized via templates) to isolate the effect of telemetry on learning and to keep linguistic realization out of scope. A natural extension is to retain DT unchanged as the measurement backbone while replacing the template actuator with richer realization mechanisms (e.g., retrieval-grounded prompting or LLM-based generation), using PE/SI as feedback signals for monitoring and optimization.

\begin{table}[]
\centering
\caption{Comparison Results - During Training (Averaged)}
\label{tab1:training_comparison_across}
\footnotesize
\begin{tabular}{@{}lccc@{}}
\toprule
\textbf{Method} & \textbf{SI} & \textbf{Total Knowledge} & \textbf{Complete Categories} \\
\midrule
\textbf{Condition A} & & & \\
Full-DT & \cellcolor{blue!20}0.009$_{0.001}$ & \cellcolor{red!20}0.76$_{0.17}$ & \cellcolor{red!20}6.51$_{2.1}$ \\
Baseline & 0.071$_{0.034}$ & 0.36$_{0.25}$ & 2.89$_{2.42}$ \\
DT (w/o SI) & \cellcolor{red!20}0.010$_{0.002}$ & \cellcolor{blue!20}0.86$_{0.009}$ & \cellcolor{blue!20}7.77$_{0.07}$ \\
\midrule
\textbf{Condition B} & & & \\
Full-DT & \cellcolor{blue!20}0.13$_{0.013}$ & \cellcolor{blue!20}0.54$_{0.08}$ & \cellcolor{blue!20}4.56$_{0.75}$ \\
Baseline & \cellcolor{red!20}0.18$_{0.028}$ & \cellcolor{red!20}0.37$_{0.06}$ & \cellcolor{red!20}2.90$_{0.54}$ \\
DT (w/o SI) & 0.22$_{0.06}$ & 0.31$_{0.055}$ & 2.33$_{0.55}$ \\
\bottomrule
\end{tabular}
\\[0.5em]
\begin{minipage}{0.9\linewidth}
\footnotesize
Reported values are $\text{mean}_{\text{std}}$. Blue cells indicate best results, red cells second-best results. The table compares agents' performance under Conditions A and B across three evaluation metrics: SI, Total Knowledge, and Complete Categories.
\end{minipage}
\end{table}

\begin{table}[]
\centering
\caption{Comparison Results - End of Training (Averaged)}
\label{tab2:training_comparison_end}
\footnotesize
\begin{tabular}{@{}lccc@{}}
\toprule
\textbf{Method} & \textbf{SI} & \textbf{Total Knowledge} & \textbf{Complete Categories} \\
\midrule
\textbf{Condition A} & & & \\
Full-DT & \cellcolor{blue!20}0.001$_{0.007}$ & \cellcolor{red!20}0.76$_{0.22}$ & \cellcolor{red!20}6.60$_{2.62}$ \\
Baseline & 0.067$_{0.08}$ & 0.22$_{0.34}$ & 1.88$_{3.13}$ \\
DT (w/o SI) & \cellcolor{blue!20}0.001$_{0.007}$ & \cellcolor{blue!20}0.87$_{0.009}$ & \cellcolor{blue!20}8.0$_{0.00}$ \\
\midrule
\textbf{Condition B} & & & \\
Full-DT & \cellcolor{blue!20}0.001$_{0.007}$ & \cellcolor{blue!20}0.79$_{0.2}$ & \cellcolor{blue!20}7.16$_{1.95}$ \\
Baseline & \cellcolor{red!20}0.18$_{0.05}$ & \cellcolor{red!20}0.40$_{0.13}$ & \cellcolor{red!20}3.24$_{1.21}$ \\
DT (w/o SI) & 0.21$_{0.08}$ & 0.32$_{0.09}$ & 2.44$_{0.80}$ \\
\bottomrule
\end{tabular}
\\[0.5em]
\begin{minipage}{0.9\linewidth}
\footnotesize
Reported values are $\text{mean}_{\text{std}}$. Blue cells indicate best results, red cells second-best results. The table compares agents' performance under Conditions A and B across three evaluation metrics: SI, Total Knowledge, and Complete Categories.
\end{minipage}
\end{table}

\subsection{Limitations and Scope}
\textbf{Controlled evaluation setting.} Our evaluation is conducted in a single SAR-inspired domain using controlled interview simulations with LLM-generated content. This enables systematic manipulation of stalling scenarios but limits claims about behavior with real human interviewees, who exhibit additional variability and context-dependent dynamics not fully captured by prompted LLMs.

\textbf{Question realization vs.\ strategy selection.} In Exp.~2, the policy selects discrete strategy--category actions that are realized via templates. This prioritizes interpretability and isolates telemetry effects but constrains linguistic naturalness and contextual adaptation. The present experiments therefore evaluate learning of \emph{what to probe} and \emph{when}, rather than \emph{how to phrase} questions.

\textbf{Pattern detection vs.\ causal diagnosis.} SI detects interaction patterns (repeated probing + semantic similarity + minimal knowledge gain) but does not diagnose why such patterns occur. In deployment, distinguishing ineffective questioning strategies from interviewee capacity constraints, environmental disruptions, or technical issues may require additional context (operator judgment, auxiliary sensors, historical patterns). DT provides measurement whereas interpretation and intervention strategy remain controller responsibilities.

\textbf{Parameter Sensitivity and Domain Adaptation.} Crucially, the DT framework is not a rigid black box. Its sensitivity to conversational cycling is modulated by tunable hyperparameters such as the gain dampening strength (\(\lambda\)), window size (\(W\)), and repetition threshold (\(r_{min}\)). This parameterized design allows the telemetry to be calibrated to the operational patience and redundancy tolerance of different domains. For instance, setting a higher dampening factor or larger repetition threshold could allow the system to accommodate high-stakes contexts (e.g., forensic interrogation) where repetitive drill-down questioning may be a valid strategy for extracting fine-grained details. Conversely, stricter settings can enforce rapid coverage in time-critical search operations. Thus, the distinction between productive thoroughness and unproductive stalling is not hard-coded, but is defined relative to these configurable domain constraints.

\textbf{Protocol alignment.} The agents are not trained to replicate expert cognitive interviewing protocols beyond the incentives encoded in our reward terms. Our results should not be read as validation of psychological appropriateness or interview quality against practitioner standards.

\textbf{Schema dependence.} DT assumes a predefined task schema (categories, weights, and any dependencies). Extending DT to settings where the schema is incomplete, evolving, or unknown would require additional mechanisms for schema construction or refinement.

The integration with RL policies illustrates DT's practical utility for automated control, with systematic performance improvements and context-dependent benefits from stall penalties. More broadly, DT provides a paradigm for instrumenting schema-grounded information acquisition with online, interpretable observables that support monitoring, diagnosis, and closed-loop policy design.

Future work should explore: (1) multi-modal information gathering that allocates effort across linguistic queries and sensor-based evidence using unified information-gain measures, (2) adaptive controllers that use telemetry trends (not only instantaneous values) to anticipate stalling and trigger earlier strategy changes, (3) human-in-the-loop deployments where DT supports dashboard-style monitoring and operator intervention, (4) extensions to partially specified or evolving schemas, and (5) richer question-realization layers while retaining DT as the measurement backbone for monitoring and optimization.

\section{Conclusion}
We propose Dialogue Telemetry (DT), a lightweight, model-agnostic turn-level measurement framework that operates over a predefined task schema. DT exposes two signals after each question--answer turn, a Progress Estimator (PE) that quantifies residual information and expected gain, and a Stalling Index (SI) that detects throughput degradation through observable interaction patterns (repeated probing + semantic similarity + low marginal gain). Across controlled SAR-inspired interviews using LLM-generated simulations, DT distinguishes efficient dialogue segments from injected stalling patterns, and RL integration illustrates how DT signals improve information acquisition when stalling carries operational costs. These results support DT as a practical instrumentation layer for schema-grounded information-gathering dialogues, providing interpretable observables for monitoring, diagnosis, and closed-loop policy design.

\bibliography{thebibliography}
\bibliographystyle{IEEEtran}

\appendix
\begin{footnotesize} 
\section{Appendix}

\subsection{Knowledge Dimensions and Importance Weights}

\begin{Verbatim}[frame=single, fontsize=\scriptsize]
KNOWLEDGE_DIMENSIONS = [
    "location",      # Last known location
    "time",          # When last seen
    "description",   # Physical appearance, clothing
    "equipment",     # Gear they had
    "medical",       # Medical conditions, needs
    "weather",       # Weather conditions
    "companions",    # Who they were with
    "intentions"     # Planned route/destination
]

DIMENSION_IMPORTANCE = {
    "location": 0.9,     # Most critical for search
    "time": 0.8,         # Important for search radius
    "description": 0.7,  # Helps identify person
    "equipment": 0.6,    # Indicates survival capability
    "medical": 0.8,      # Critical for urgency
    "weather": 0.5,      # Context for conditions
    "companions": 0.6,   # Potential witnesses
    "intentions": 0.7    # Direction of travel
}

DIMENSION_DEPENDENCIES = {
    "intentions": ["location"], # Route matters 
    more with starting point
    "companions": ["description"] # Companion details
    matter more with ID
}
\end{Verbatim}

\subsection{Question Strategies}
Six interviewing strategies adapted from cognitive interview practice:

\begin{description}[style=nextline,leftmargin=1.0cm,labelindent=0.5cm]

  \item[FREE\_RECALL] 
  Open-ended information gathering. \\
  \textit{Example:} ``Can you please tell me everything you remember about the missing hiker?''

  \item[SPECIFIC\_DETAIL] 
  Direct question about dimension. \\
  \textit{Examples:}
  \begin{itemize}[noitemsep,topsep=0pt,leftmargin=*]
      \item ``Where exactly was the hiker last seen?''
      \item ``What equipment or supplies did they have with them?''
  \end{itemize}

  \item[CLARIFICATION] 
  Resolve ambiguity in previous answer. \\
  \textit{Example:} ``You mentioned they were at Pine Ridge trailhead. Can you be more specific about that location?''

  \item[CONTEXT\_REINSTATE] 
  Mental recreation of situation. \\
  \textit{Example:} ``Try to put yourself back at the moment you last saw them. Look around in your mind. What landmarks or features do you see?''

  \item[SUMMARY\_CONFIRM] 
  Verify understanding. \\
  \textit{Example:} ``So to confirm what we know: the hiker was last seen at Pine Ridge trailhead around 2:17 PM, wearing a red North Face jacket\ldots''

  \item[RAPPORT\_BUILD] 
  Build trust. \\
  \textit{Example:} ``I understand this is a stressful situation. You're doing really well helping us gather information. How are you holding up?''

\end{description}

\begin{Verbatim}[frame=single, fontsize=\scriptsize]
PREDEFINED_QUESTIONS = {
    # Free recall
    ("FREE_RECALL", "general"): 
        "Can you please tell me everything you remember 
        about the missing hiker?",
    
    # Specific details
    ("SPECIFIC_DETAIL", "location"): 
        "Where exactly was the hiker last seen?",
    ("SPECIFIC_DETAIL", "time"): 
        "When exactly did you last see or hear from them?",
    ("SPECIFIC_DETAIL", "medical"): 
        "Does the hiker have any medical conditions we 
        should know about?",
    
    # Clarifications (with placeholders)
    ("CLARIFICATION", "location"): 
        "You mentioned they were at {0}. Can you be more 
        specific about that location?",
    ("CLARIFICATION", "time"): 
        "To clarify, you said around {0}. Can you narrow 
        down that timeframe?",
    
    # Context reinstatement
    ("CONTEXT_REINSTATE", "location"): 
        "Try to put yourself back at the moment you last
        saw them. Look around in your mind. 
        What landmarks or features do you see?",
    
    # ... [full template in repository]
}
\end{Verbatim}

\subsection{LLM-Based Knowledge Scoring -- Prompt}

\begin{lstlisting}[caption={LLM Knowledge Scoring Prompt}, label={lst:prompt}]
You are an expert first responder conducting on-the-ground interviews about a missing hiker. You will:

1. Analyze the given text for each dimension tailored to the missing-hiker scenario.
2. Extract the specific nugget(s) of information for that dimension.
3. Score each dimension from 0.0 to 1.0 based on how actionable and precise the information is.
4. Return ONLY a JSON object with scores for all dimensions.

DIMENSIONS TO ANALYZE:
- LOCATION: Geographic information, places, addresses, directions
- TIME: Temporal information, dates, times, durations
- MEDICAL: Health conditions, symptoms, medications, treatments
- EQUIPMENT: Tools, devices, gear, technology mentioned
- DESCRIPTION: Physical appearance, clothing, identifying features
- WEATHER: Weather conditions, forecasts, environmental factors
- COMPANIONS: Who they were with, group information, other people
- INTENTIONS: Planned destination, route, purpose, activities

SCORING SCALE (0.0-1.0):
0.9-1.0: Highly specific, explicit, actionable information about the hiker's current situation.
0.7-0.8: Moderately specific, clear, relevant information (e.g., named trail segments, landmarks).
0.5-0.6: General information that may help narrow search area but lacks detail (e.g., "in the forest").
0.3-0.4: Vague or implied info not precisely locatable (e.g., "near a road").
0.1-0.2: Barely detectable hints (e.g., "he might have left home").
0.0: No information relevant to this dimension.

HOW TO SCORE "LOCATION" (example):
- A mention of "his house" scores lower because it's static/past and doesn't help you locate him now.
- A mention of "Pine Ridge trail, halfway up the summit path" scores high because it's specific and actionable.
- If both appear, weight the more recent, dynamic, terrain-specific clue.

INSTRUCTIONS:
1. For each dimension, ask yourself:
   - Relevance: Does this help me search right now?
   - Specificity: Are there named trails, landmarks, compass directions?
   - Recency: Is it a "last known" sighting, or something too general/past?
2. Extract the exact phrase(s) that justify your score.
3. Provide a one-sentence reasoning citing that phrase.

TEXT TO ANALYZE:
"{text}"

REQUIRED OUTPUT FORMAT:
{
  "location":   { "score": 0.0, "extracted_info": "", "reasoning": "" },
  "time":       { "score": 0.0, "extracted_info": "", "reasoning": "" },
  "medical":    { "score": 0.0, "extracted_info": "", "reasoning": "" },
  "equipment":  { "score": 0.0, "extracted_info": "", "reasoning": "" },
  "description":{ "score": 0.0, "extracted_info": "", "reasoning": "" },
  "weather":    { "score": 0.0, "extracted_info": "", "reasoning": "" },
  "companions": { "score": 0.0, "extracted_info": "", "reasoning": "" },
  "intentions": { "score": 0.0, "extracted_info": "", "reasoning": "" }
}

JSON OUTPUT ONLY.
\end{lstlisting}

\begin{figure*}[]
\centering
\includegraphics[width=\textwidth]{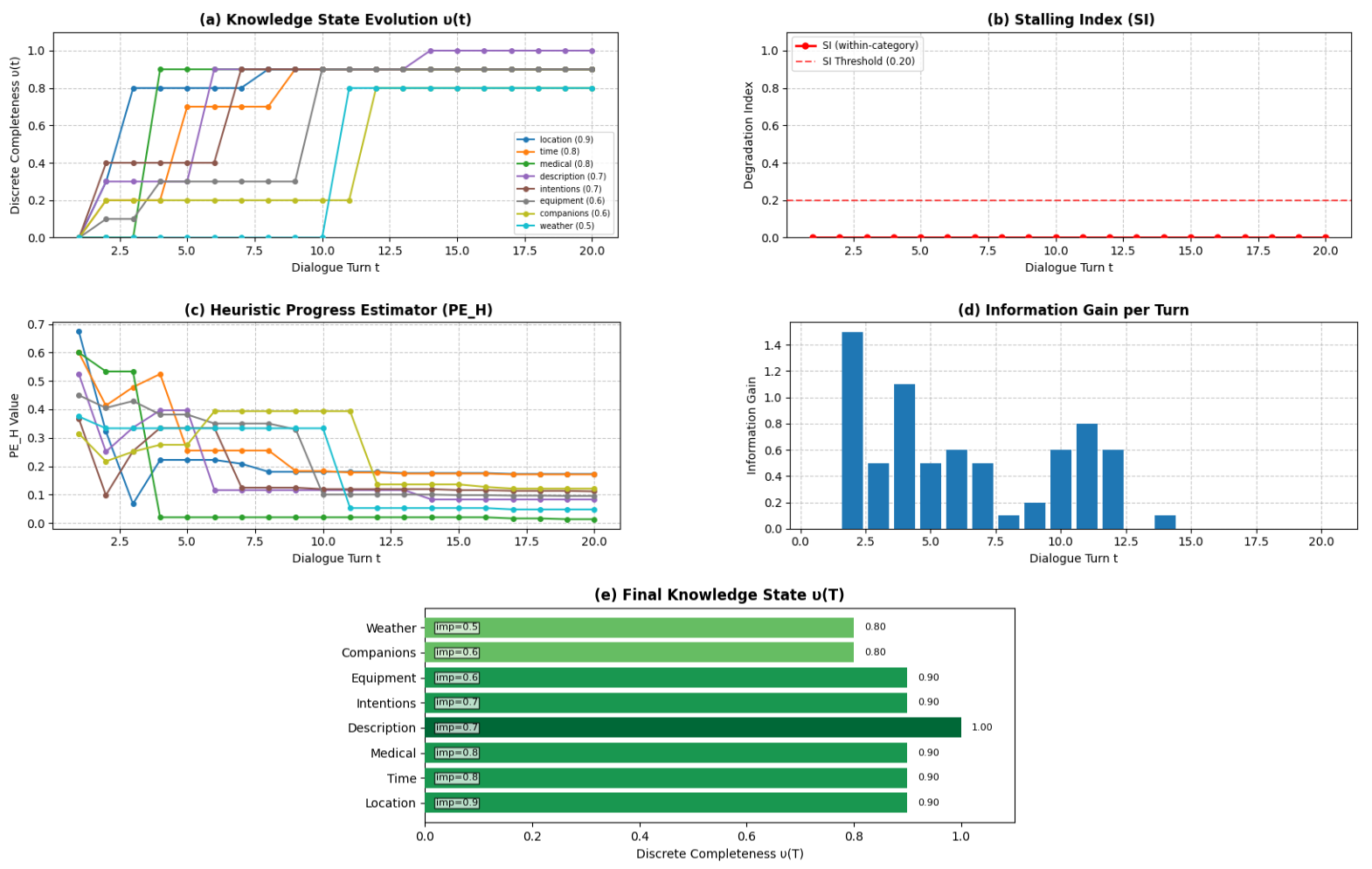}
\caption{DT Telemetry Monitoring of Efficient Dialogue Using Heuristic PE Variant. Turn-by-turn tracking of optimal 20-turn \(\mathrm{DT}\)-guided interrogation using \(\mathrm{PE}^{\text{H}}\) (Eq.~\ref{eq4:heuristic_PE}). (a) Systematic knowledge evolution achieving high completion rates; (b) Stall Index (\(\mathrm{SI}\)) remaining below detection threshold throughout; (c) Expected gain (heuristic) values decreasing systematically as gaps close; (d) Information gain following diminishing returns pattern; (e) Final state showing successful completion aligned with importance weights. Compare with Fig.~\ref{fig2:results_4-2-efficient} which uses the formal entropy-based PE variant.}
\label{fig22:results_4-2-efficient-heuristic-APPENDIX}
\end{figure*}

\subsection*{Scenario 1: Efficient Dialogue Transcript}
\textit{Characteristics: Systematic progression across high-importance categories, minimal repetition, high information yield per turn, SI remains below detection threshold throughout.}

\vspace{0.3cm}

\textbf{Turn 1} | RAPPORT\_BUILD $\rightarrow$ general \\
\textbf{Q}: I understand this is a stressful situation. You're doing really well helping us gather information. How are you holding up? \\
\textbf{A}: I'm really worried about him. He's usually very reliable and always lets me know if his plans change. I didn't sleep at all last night when I realized he hadn't checked in. Thank you for taking this seriously. I just want to make sure he's okay.

\vspace{0.2cm}

\textbf{Turn 2} | FREE\_RECALL $\rightarrow$ general \\
\textbf{Q}: Can you please tell me everything you remember about the missing hiker? \\
\textbf{A}: His name is Tom Richardson, about 35 years old. He was heading to Pine Ridge trail yesterday. He was wearing something red, I think. He drove himself to the trailhead and was planning a day hike, not overnight. He's an experienced hiker but this is his first time on this particular trail. 

\vspace{0.2cm}

\textbf{Turn 3} | SPECIFIC\_DETAIL $\rightarrow$ location \\
\textbf{Q}: Where exactly was the hiker last seen? \\
\textbf{A}: At the Pine Ridge trailhead parking lot, near the eastern entrance of the national forest. That's where his car is still parked.

\vspace{0.2cm}

\textbf{Turn 4} | SPECIFIC\_DETAIL $\rightarrow$ medical \\
\textbf{Q}: Does the hiker have any medical conditions we should know about? \\
\textbf{A}: Yes, he's diabetic and needs insulin daily. He would have brought his medication with him, but I'm not sure how much. He also has a mild allergy to bee stings - not life-threatening but something to be aware of.

\vspace{0.2cm}

\textbf{Turn 5} | SPECIFIC\_DETAIL $\rightarrow$ time \\
\textbf{Q}: When exactly did you last see or hear from them? \\
\textbf{A}: Around 2 PM yesterday afternoon. He texted me when he arrived at the trailhead.

\vspace{0.2cm}

\textbf{Turn 6} | SPECIFIC\_DETAIL $\rightarrow$ description \\
\textbf{Q}: Can you describe in detail what the hiker was wearing and what they look like? \\
\textbf{A}: Tom is about 6 feet tall, athletic build, with short brown hair and a short beard. He was wearing a bright red North Face fleece jacket, tan hiking pants, and a black backpack. He also had a blue baseball cap with a mountain logo on it. He wears glasses for distance.

\vspace{0.2cm}

\textbf{Turn 7} | SPECIFIC\_DETAIL $\rightarrow$ intentions \\
\textbf{Q}: Did they mention where they were planning to go or their route? \\
\textbf{A}: He was planning to hike the Pine Ridge loop trail, which is about 8 miles total. He mentioned wanting to reach the summit viewpoint, which is about halfway around the loop. He was going to hike clockwise around the loop.

\vspace{0.2cm}

\textbf{Turn 8} | CLARIFICATION $\rightarrow$ location \\
\textbf{Q}: You mentioned they were at Pine Ridge trailhead. Can you be more specific about that location? \\
\textbf{A}: The eastern trailhead has a split about half a mile in. He was planning to take the right fork toward the ridge. The parking lot is just off Highway 16, about 5 miles after the forest entrance station. There's a large wooden sign and a small ranger station that's usually unmanned.

\vspace{0.2cm}

\textbf{Turn 9} | CLARIFICATION $\rightarrow$ time \\
\textbf{Q}: To clarify, you said around 2 PM. Can you narrow down that timeframe? \\
\textbf{A}: I received a text from him at exactly 2:17 PM saying he had reached the trailhead. He sent a photo of the trail map at 2:22 PM. That was the last I heard from him.

\vspace{0.2cm}

\textbf{Turn 10} | SPECIFIC\_DETAIL $\rightarrow$ equipment \\
\textbf{Q}: What equipment or supplies did they have with them? \\
\textbf{A}: He had a black backpack, about 30 liters I think. He was carrying water bottles, some trail mix and energy bars, and a small first aid kit. He had his phone with him, but the battery doesn't last long. I think he had a portable charger. He also had a printed map of the trail system. No tent or sleeping bag since it was supposed to be a day hike.

\vspace{0.2cm}

\textbf{Turn 11} | SPECIFIC\_DETAIL $\rightarrow$ weather \\
\textbf{Q}: What were the weather conditions in the area? \\
\textbf{A}: It was sunny and clear when he left, around 65 degrees. But I checked the forecast later and there were thunderstorms predicted for the evening. The temperature was supposed to drop to around 40 overnight. I don't know if it actually rained in that area though.

\vspace{0.2cm}

\textbf{Turn 12} | SPECIFIC\_DETAIL $\rightarrow$ companions \\
\textbf{Q}: Was the hiker with anyone else? If so, who? \\
\textbf{A}: No, he was hiking alone. He usually goes with his friend David, but David couldn't make it yesterday.

\vspace{0.2cm}

\textbf{Turn 13} | CONTEXT\_REINSTATE $\rightarrow$ location \\
\textbf{Q}: Try to put yourself back at the moment you last saw them. Look around in your mind. What landmarks or features do you see? \\
\textbf{A}: I didn't see him at the trailhead - we just texted. But from the photo he sent me, I can see the large trailhead sign with the map display. There's a wooden footbridge visible just beyond the sign where the trail begins. The parking lot has a gravel surface, not paved. There were about 5-6 other cars in the parking lot in his photo.

\vspace{0.2cm}

\textbf{Turn 14} | CONTEXT\_REINSTATE $\rightarrow$ description \\
\textbf{Q}: Close your eyes for a moment and picture the hiker. Starting from their head down to their feet, what details can you recall? \\
\textbf{A}: He has short brown hair, a bit wavy on top, with a neatly trimmed beard. He wears rectangular-framed glasses. He had his blue cap with the mountain logo. The red fleece jacket was zipped up when I saw him. Underneath he usually wears a moisture-wicking t-shirt, probably gray or black. Tan hiking pants with zippered pockets. He was wearing his hiking boots - they're brown leather Merrells that he's had for years. He also wears a black fitness watch - I think it's a Garmin.

\vspace{0.2cm}

\textbf{Turn 15} | SPECIFIC\_DETAIL $\rightarrow$ equipment \\
\textbf{Q}: You said they had a printed map. Do you know specifically what kind or how much? \\
\textbf{A}: It was the official trail map from the forest service. He printed it from their website the night before. It shows all the trails in the Pine Ridge area, including emergency exit routes and water sources. 

\vspace{0.2cm}

\textbf{Turn 16} | SPECIFIC\_DETAIL $\rightarrow$ companions \\
\textbf{Q}: You said they were hiking alone. Do you have any contact information for his friend David? \\
\textbf{A}: Yes, David Wilson is his usual hiking partner. I have his phone number: 555-123-4567. He might know more about Tom's hiking habits and abilities.

\vspace{0.2cm}

\textbf{Turn 17} | FREE\_RECALL $\rightarrow$ general \\
\textbf{Q}: Can you please tell me everything you remember about the missing hiker? \\
\textbf{A}: His name is Tom Richardson, about 35 years old. He was heading to Pine Ridge trail yesterday. He was wearing something red, I think. He drove himself to the trailhead and was planning a day hike, not overnight. He's an experienced hiker but this is his first time on this particular trail.

\vspace{0.2cm}

\textbf{Turn 18} | CLARIFICATION $\rightarrow$ general \\
\textbf{Q}: You mentioned red fleece jacket. Can you describe that in more detail? \\
\textbf{A}: It's a bright red North Face fleece full-zip jacket, fairly new. It has black trim around the zippers and a small North Face logo on the chest. It's not a heavy winter jacket, more of a mid-layer.

\vspace{0.2cm}

\textbf{Turn 19} | CLARIFICATION $\rightarrow$ general \\
\textbf{Q}: Regarding the diabetes you mentioned, do you know if they had medication or any specific needs? \\
\textbf{A}: He uses an insulin pen that he would keep in an insulated pouch in his backpack. He always carries glucose tablets for emergencies. He needs to check his blood sugar regularly, especially when exercising. 

\vspace{0.2cm}

\textbf{Turn 20} | SUMMARY\_CONFIRM $\rightarrow$ general \\
\textbf{Q}: So to confirm what we know: the hiker was last seen at Pine Ridge trailhead around 2:17 PM, wearing a red North Face jacket and tan pants. They had hiking supplies with them, and were planning to hike the loop clockwise. Is all that correct? \\
\textbf{A}: Yes, that's all correct. The only thing I'd add is that he was planning to reach the summit viewpoint and might have taken the Cedar Creek shortcut if the weather turned bad.

\begin{figure*}[]
\centering
\includegraphics[width=\textwidth]{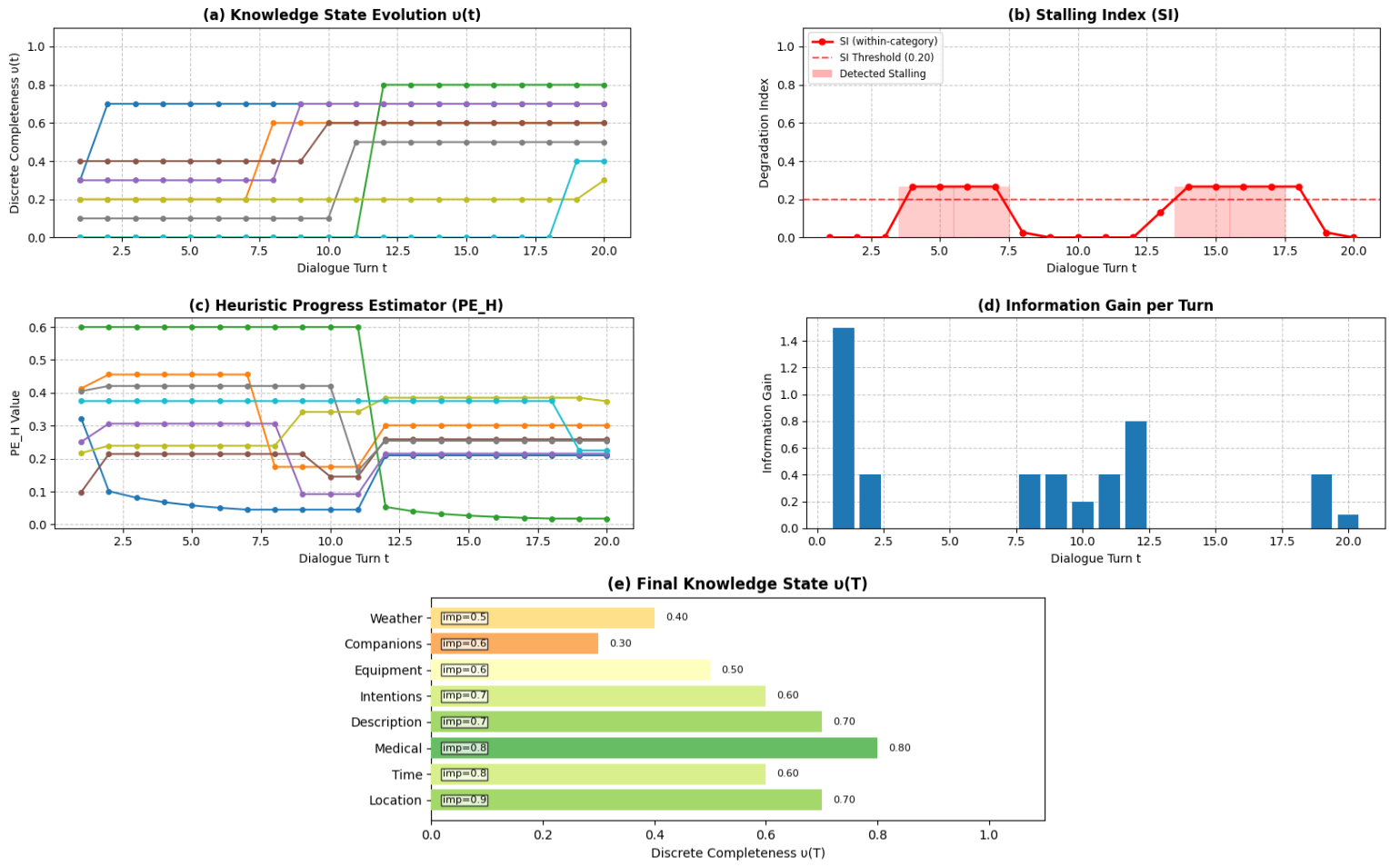}
\caption{DT Telemetry Detection of Conversational Stalling Using Heuristic PE Variant. Monitoring of a 20-turn dialogue exhibiting dual stalling incidents using \(\mathrm{PE}^{\text{H}}\) (Eq.~\ref{eq4:heuristic_PE}). (a) Knowledge evolution showing suboptimal progression; (b) \(\mathrm{SI}\) detecting two stalling periods (red shaded) exceeding threshold; (c) Expected gain (heuristic) values remaining persistently high due to unresolved information gaps; (d) Information gain showing near-zero acquisition during stalling periods; (e) Final state revealing poor completion across categories. Compare with Fig.~\ref{fig3:dual} which uses the formal entropy-based PE variant.}
\label{fig33:dual-heuristic-APPENDIX}
\end{figure*}

\subsection*{Scenario 2: Agent-Side Repetition Transcript}
\textit{Characteristics: Agent repeatedly probes same categories with similar questions, resulting in diminishing information yield. Two stalling episodes occur (Turns 2-7 on location, Turns 12-18 on medical), with SI detecting both.}

\vspace{0.3cm}

\textbf{Turn 1} | FREE\_RECALL $\rightarrow$ general \\
\textbf{Q}: Can you please tell me everything you remember about the missing hiker? \\
\textbf{A}: Okay, um, his name's Tom Richardson. He's 35, I think? Yeah, 35. He went to Pine Ridge yesterday morning. I remember he was wearing something red - a jacket or fleece, bright red. He drove his own car there, you know, to the trailhead. He said it was just gonna be a day hike, nothing overnight. Tom's done a lot of hiking before, but... but I don't think he's been on this trail specifically.

\vspace{0.2cm}

\noindent\colorbox{red!20}{\parbox{\columnwidth}{\textbf{STALL EPISODE 1: Location Dimension (Turns 2-7)}}}

\textbf{Turn 2} | SPECIFIC\_DETAIL $\rightarrow$ location \\
\textbf{Q}: Where exactly was the hiker last seen? \\
\textbf{A}: He was at the Pine Ridge trailhead. The parking lot there, near the... I think it's the eastern entrance? Yeah, eastern side of the national forest. His car's still there. I saw it when I drove by looking for him. \\
\textit{[Knowledge gain: location +0.70 -- good initial information]}

\vspace{0.2cm}

\textbf{Turn 3} | CLARIFICATION $\rightarrow$ location \\
\textbf{Q}: Can you be more specific about the Pine Ridge trailhead location? \\
\textbf{A}: It's at the eastern entrance like I said. There's a parking lot there, I think maybe 20 spaces or so? \\
\textit{[Knowledge gain: location +0.05 -- repetitive, minimal new information]}

\vspace{0.2cm}

\textbf{Turn 4} | SPECIFIC\_DETAIL $\rightarrow$ location \\
\textbf{Q}: Are there any landmarks or features near the trailhead that could help us? \\
\textbf{A}: Well, it's near the national forest entrance. I think there might be some signs? But I'm not sure what else is there. \\
\textit{[Knowledge gain: location +0.03 -- vague, almost no gain]}

\vspace{0.2cm}

\textbf{Turn 5} | SPECIFIC\_DETAIL $\rightarrow$ location \\
\textbf{Q}: You mentioned the eastern entrance. Can you describe the road or access to get there? \\
\textbf{A}: I don't really know the details of how to get there. Like, I just know it's the eastern entrance to the forest. \\
\textit{[Knowledge gain: location +0.01 -- SI begins detecting stalling]}

\vspace{0.2cm}

\textbf{Turn 6} | SPECIFIC\_DETAIL $\rightarrow$ location \\
\textbf{Q}: What's the exact address or GPS coordinates of the trailhead? \\
\textbf{A}: I don't know the exact address. It's just... it's the Pine Ridge trailhead at the eastern forest entrance. \\
\textit{[Knowledge gain: location +0.0 -- no new information, SI elevated]}

\vspace{0.2cm}

\textbf{Turn 7} | SPECIFIC\_DETAIL $\rightarrow$ location \\
\textbf{Q}: Are there any other parking areas or access points nearby? \\
\textbf{A}: I think that's the main parking area. I'm not really familiar with other access points. \\
\textit{[Knowledge gain: location +0.0 -- stalling episode continues]}

\vspace{0.2cm}


\textbf{Turn 8} | SPECIFIC\_DETAIL $\rightarrow$ time \\
\textbf{Q}: When exactly did you last see or hear from them? \\
\textbf{A}: It was yesterday afternoon. Around 2, maybe a little after? He texted me when he got there - I remember looking at my phone and it was like 2:15 or so. That was the last time I heard from him. \\
\textit{[Knowledge gain: time +0.60 -- recovery, productive turn]}

\vspace{0.2cm}

\textbf{Turn 9} | SPECIFIC\_DETAIL $\rightarrow$ description \\
\textbf{Q}: Can you describe in detail what the hiker was wearing and what they look like? \\
\textbf{A}: Tom's about 6 feet tall, maybe a bit under. He's in good shape, you know, athletic. He's got short brown hair and this beard - not long, just like a short beard. Oh, and he wears glasses for seeing far away. Um, he was definitely wearing that bright red fleece jacket - North Face I think? And tan pants, hiking pants. \\
\textit{[Knowledge gain: description +0.70 -- informative]}

\vspace{0.2cm}

\textbf{Turn 10} | SPECIFIC\_DETAIL $\rightarrow$ intentions \\
\textbf{Q}: Did they mention where they were planning to go or their route? \\
\textbf{A}: He said he was doing the Pine Ridge loop trail. It's like 8 miles around? He wanted to get to this viewpoint at the top - the summit viewpoint. \\
\textit{[Knowledge gain: intentions +0.60 -- useful information]}

\vspace{0.2cm}

\textbf{Turn 11} | SPECIFIC\_DETAIL $\rightarrow$ equipment \\
\textbf{Q}: What equipment or supplies did they have with them? \\
\textbf{A}: He had his black backpack and was carrying water bottles and some snacks. His phone was with him. \\
\textit{[Knowledge gain: equipment +0.50 -- basic information]}

\vspace{0.2cm}

\noindent\colorbox{red!20}{\parbox{\columnwidth}{\textbf{STALL EPISODE 2: Medical Dimension (Turns 12-18)}}}

\textbf{Turn 12} | SPECIFIC\_DETAIL $\rightarrow$ medical \\
\textbf{Q}: Does the hiker have any medical conditions we should know about? \\
\textbf{A}: Yes! That's what I'm really worried about. He's diabetic - he needs insulin every day. He would've packed his medication but I don't know how much he brought. And he's allergic to bee stings. Not super serious, like he's not gonna die from it, but it's something you should know about. \\
\textit{[Knowledge gain: medical +0.80 -- highly informative baseline]}

\vspace{0.2cm}

\textbf{Turn 13} | SPECIFIC\_DETAIL $\rightarrow$ medical \\
\textbf{Q}: Can you tell me more about his diabetes management? \\
\textbf{A}: He uses insulin, like I said. He's pretty good about managing it, you know? \\
\textit{[Knowledge gain: medical +0.02 -- repetitive, minimal gain]}

\vspace{0.2cm}

\textbf{Turn 14} | SPECIFIC\_DETAIL $\rightarrow$ medical \\
\textbf{Q}: What type of insulin does he use? Pen or vial? \\
\textbf{A}: I think it's a pen? But I'm not really sure about the specifics of his medication. \\
\textit{[Knowledge gain: medical +0.01 -- uncertain, minimal information]}

\vspace{0.2cm}

\textbf{Turn 15} | SPECIFIC\_DETAIL $\rightarrow$ medical \\
\textbf{Q}: How often does he need to take his insulin? \\
\textbf{A}: Daily, like I mentioned. I don't know the exact schedule. \\
\textit{[Knowledge gain: medical +0.005 -- SI begins detecting second stalling episode]}

\vspace{0.2cm}

\textbf{Turn 16} | SPECIFIC\_DETAIL $\rightarrow$ medical \\
\textbf{Q}: Does he carry any emergency glucose or medical supplies for his diabetes? \\
\textbf{A}: I assume so, since he's diabetic, right? But I don't know exactly what he carries with him. \\
\textit{[Knowledge gain: medical +0.0 -- no new information]}

\vspace{0.2cm}

\textbf{Turn 17} | SPECIFIC\_DETAIL $\rightarrow$ medical \\
\textbf{Q}: How severe is his bee sting allergy? Does he carry an EpiPen? \\
\textbf{A}: It's mild, like I said. I don't think he carries an EpiPen or anything. \\
\textit{[Knowledge gain: medical +0.005 -- SI elevated]}

\vspace{0.2cm}

\textbf{Turn 18} | SPECIFIC\_DETAIL $\rightarrow$ medical \\
\textbf{Q}: Are there any other medical issues we should know about? \\
\textbf{A}: No, just the diabetes and bee sting allergy I already mentioned. \\
\textit{[Knowledge gain: medical +0.0 -- stalling episode continues]}

\vspace{0.2cm}


\textbf{Turn 19} | SPECIFIC\_DETAIL $\rightarrow$ weather \\
\textbf{Q}: What were the weather conditions in the area? \\
\textbf{A}: When he left it was really nice. Sunny, clear. But I heard there might be storms later that evening? \\
\textit{[Knowledge gain: weather +0.40 -- incomplete information due to time pressure]}

\vspace{0.2cm}

\textbf{Turn 20} | SPECIFIC\_DETAIL $\rightarrow$ companions \\
\textbf{Q}: Was the hiker with anyone else? \\
\textbf{A}: No, no one. He was by himself. \\
\textit{[Knowledge gain: companions +0.30 -- insufficient coverage due to wasted turns]}

\vspace{0.5cm}

\vspace{0.3cm}

\noindent\textit{Note: Turns 2-7 demonstrate agent circling inefficiently on location dimension with 6 consecutive questions yielding minimal incremental knowledge (0.70 → 0.05 → 0.03 → 0.01 → 0.0 → 0.0). Similarly, Turns 12-18 show repetitive medical probing with 7 consecutive questions producing diminishing returns (0.80 → 0.02 → 0.01 → 0.005 → 0.0 → 0.005 → 0.0). This agent-side repetition contrasts with Scenario 3 where questions are varied but responses degrade, demonstrating SI's cause-agnostic detection capability.}

\end{footnotesize}
\end{document}